\documentclass{IEEEtran}
\usepackage{amssymb}
\usepackage{graphicx}
\usepackage{colortbl}
\usepackage{cite}
\usepackage{booktabs}
\usepackage{multirow}
\usepackage{multicol}
\usepackage{amsfonts}
\usepackage{textcomp}
\usepackage[hidelinks]{hyperref}
\usepackage{changepage}
\usepackage{algorithm} 
\usepackage{algorithmic}
\definecolor{Asphalt}{RGB}{255,0,0}
\definecolor{Meadows}{RGB}{230,0,255}
\definecolor{Gravel}{RGB}{164,75,155}
\definecolor{Trees}{RGB}{255,255,0}
\definecolor{Metal sheets}{RGB}{0,110,0}
\definecolor{Bare soil}{RGB}{102,255,255}
\definecolor{Bitumen}{RGB}{140,67,46}
\definecolor{Bricks}{RGB}{200,200,200}
\definecolor{Shadows}{RGB}{101,193,60}

\definecolor{Roof}{RGB}{204,102,102}
\definecolor{Grass}{RGB}{153, 51, 0}
\definecolor{Road}{RGB}{204, 153, 0}
\definecolor{Train}{RGB}{0, 255, 0}
\definecolor{Tree}{RGB}{0,102,0}
\definecolor{Shadow}{RGB}{0,51,255}

\definecolor{Grass-healthy}{RGB}{0,205,0}
\definecolor{Grass-stressed}{RGB}{127, 255, 0}
\definecolor{Grass-synthetic}{RGB}{46, 139, 87}
\definecolor{Tree}{RGB}{0, 139, 0}
\definecolor{Soil}{RGB}{160, 82, 45}
\definecolor{Water}{RGB}{0,255,255}
\definecolor{Residential}{RGB}{255,255,255}
\definecolor{Commercial}{RGB}{216, 191, 216}
\definecolor{Road}{RGB}{255, 0, 0}
\definecolor{Highway}{RGB}{139, 0, 0}
\definecolor{Railway}{RGB}{0,0,0}
\definecolor{Parking-lot1}{RGB}{255,255,0}
\definecolor{Parking-lot2}{RGB}{238, 154, 0}
\definecolor{Tennis-court}{RGB}{85, 26, 139}
\definecolor{Running-track}{RGB}{255, 127, 80}

\definecolor{Corn-high}{RGB}{140,67,46}
\definecolor{Corn-mid}{RGB}{0,0,255}
\definecolor{Corn-low}{RGB}{255,100,0}
\definecolor{Soy-bean-high}{RGB}{0,255,123}
\definecolor{Soy-bean-mid}{RGB}{164,75,155}
\definecolor{Soy-bean-low}{RGB}{101,174,255}
\definecolor{Residues}{RGB}{118,254,172}
\definecolor{Wheat}{RGB}{60,91,112}
\definecolor{Hay}{RGB}{255,255,0}
\definecolor{Grass/Pasture}{RGB}{255,255,125}
\definecolor{Cover Crop 1}{RGB}{255,0,255}
\definecolor{Cover Crop 2}{RGB}{100,0,255}
\definecolor{Woodlands}{RGB}{0,172,254}
\definecolor{Highway}{RGB}{0,255,0}
\definecolor{Local Road}{RGB}{171,175,80}
\definecolor{Buildings}{RGB}{101,193,60}

\definecolor{Anhydrite}{RGB}{152,251,152}
\definecolor{Quartz}{RGB}{245,222,179}
\definecolor{Sulfides}{RGB}{139,69,19}
\definecolor{Muscovite}{RGB}{34,139,34}
\definecolor{Feldspar}{RGB}{255,201,14}

\definecolor{Healthy Grass}{RGB}{0, 255, 0}
\definecolor{Stressed Grass}{RGB}{127, 255, 0}
\definecolor{Artificial Turf}{RGB}{46, 139, 87}
\definecolor{Evergreen Trees}{RGB}{0, 139, 0}
\definecolor{Deciduous Trees}{RGB}{0, 70, 0}
\definecolor{Bare Earth}{RGB}{160, 82, 45}
\definecolor{Water}{RGB}{0, 255, 255}
\definecolor{Residential }{RGB}{255, 255, 255}
\definecolor{Commercial}{RGB}{216, 191, 216}
\definecolor{Roads}{RGB}{255, 0, 0}
\definecolor{Sidewalks}{RGB}{170, 160, 150}
\definecolor{Crosswalks}{RGB}{128, 128, 128}
\definecolor{Major Thoroughfares}{RGB}{160, 0, 0}
\definecolor{Highways}{RGB}{80,0,0}
\definecolor{Railways}{RGB}{232, 161, 24}
\definecolor{Paved Parking Lots}{RGB}{255, 255, 0}
\definecolor{Gravel Parking Lots}{RGB}{238, 154, 0}
\definecolor{Cars}{RGB}{255, 0, 255}
\definecolor{Trains}{RGB}{0, 0, 255}
\definecolor{Stadium Seats}{RGB}{176, 196, 222}
\ifCLASSINFOpdf
\else
\fi
\usepackage[cmex10]{amsmath}

\usepackage{mdwmath}

\usepackage{bm}

\usepackage[tight,footnotesize,center]{subfigure}
\begin{document}
%
\title{Fusion of Dual Spatial Information for Hyperspectral Image Classification}

%
%

\author{ Puhong Duan,~\IEEEmembership{Student Member,~IEEE,}
         Pedram~Ghamisi,~\IEEEmembership{Senior Member,~IEEE,}
         Xudong~Kang,~\IEEEmembership{Senior Member,~IEEE,}
         Behnood~Rasti,~\IEEEmembership{Senior Member,~IEEE,}
         Shutao~Li,~\IEEEmembership{Fellow,~IEEE}
	    ~and~Richard~Gloaguen

\thanks{This paper is supported by the Major Program of the National Natural Science Foundation of China (No. 61890962), the National Natural Science Foundation of China (No. 61871179), the National Natural Science Fund of China for International Cooperation and Exchanges (No. 61520106001), the Fund of Key Laboratory of Visual Perception and Artificial Intelligence of Hunan Province (No. 2018TP1013), the Fund of Hunan Province for Science and Technology Plan Project under Grant (No. 2017RS3024) and the China Scholarship Council.}
\thanks{P. Duan is with the College of Electrical and Information Engineering, Hunan University, 410082 Changsha, China, and also with the Key Laboratory of Visual Perception and Artificial Intelligence of Hunan Province, Changsha, 410082, China, and also  with the Helmholtz-Zentrum Dresden-Rossendorf, Helmholtz Institute Freiberg for Resource Technology, Freiberg 09599, Germany. e-mail: (puhong\_duan@hnu.edu.cn)}
\thanks{P. Ghamisi, B. Rasti, and R. Gloaguen are with the Helmholtz-Zentrum Dresden-Rossendorf, Helmholtz Institute Freiberg for Resource Technology, Freiberg 09599, Germany. e-mail: (p.ghamisi@gmail.com; behnood.rasti@gmail.com; r.gloaguen@hzdr.de) }
\thanks{X. Kang, and S. Li are with the College of Electrical and Information Engineering, Hunan University, 410082 Changsha, China, and also with the Key Laboratory of Visual Perception and Artificial Intelligence of Hunan Province, Changsha, 410082, China. e-mail: (xudong$\_$kang@163.com; shutao\_li@hnu.edu.cn)}}



%
%

\markboth{IEEE Transactions on Geoscience and Remote Sensing}%
{Shell \MakeLowercase{\textit{et al.}}: TGRS_DAP}
%



\maketitle

\begin{abstract}
The inclusion of spatial information into spectral classifiers for fine-resolution hyperspectral imagery has led to significant improvements in terms of classification performance. The task of spectral-spatial hyperspectral image classification has remained challenging because of high intraclass spectrum variability and low interclass spectral variability. This fact has made the extraction of spatial information highly active. In this work, a novel hyperspectral image classification framework using the fusion of dual spatial information is proposed, in which the dual spatial information is built by both exploiting pre-processing feature extraction and post-processing spatial optimization. In the feature extraction stage, an adaptive texture smoothing method is proposed to construct the structural profile (SP), which makes it possible to precisely extract discriminative features from hyperspectral images. The SP extraction method is used here for the first time in the remote sensing community. Then, the extracted SP is fed into a spectral classifier. In the spatial optimization stage, a pixel-level classifier is used to obtain the class probability followed by an extended random walker-based spatial optimization technique. Finally, a decision fusion rule is utilized to fuse the class probabilities obtained by the two different stages. Experiments performed on three data sets from different scenes illustrate that the proposed method can outperform other state-of-the-art classification techniques. In addition, the proposed feature extraction method, i.e., SP, can effectively improve the discrimination between different land covers.
\end{abstract}
\section{Introduction}
\label{sec:intro}
Hyperspectral images (HSIs) can provide hundreds of continues spectral bands carrying abundant spectral information, which is helpful for identifying different materials of interest \cite{6832757,8340224,8697135}. Owing to this characteristic, HSIs have been extensively applied in classification-related tasks such as mineral mapping \cite{Acosta,rs12182903}, land cover investigation \cite{PAN2018108,Gu_Liu}, and environmental monitoring \cite{HE201765,DUAN2020359}. Hyperspectral image classification has always been regarded as a hot topic in the remote sensing community since this technology is able to provide high-level interpretation which is beneficial for management decisions \cite{DHV,PCAEPFs,Lu_RL,Lu_ML,HONG202012}. \par
Over the past years, various supervised classifiers have been successfully applied for the classification of HSIs \cite{Advanced_spec}, including support vector machine (SVM) \cite{SVM}, multinomial logistic regression (MLR) \cite{MLR}, and random forest \cite{8046025}. These approaches have achieved satisfactory results with a certain number of high-quality samples. Nevertheless, when the number of training samples is very limited, these spectral classifiers cannot work well because of the curse of dimensionality. In addition, the high spectral dimensionality of such data sets also increases the computing cost of the classification task. Accordingly, many dimension reduction methods including feature selection and manifold learning have been developed to decrease the spectral dimension of HSIs. For example, in \cite{4786582}, the linear discriminant analysis was employed to decrease the high dimensionality of the original HSIs. In \cite{SBS}, an efficient manifold ranking-based method for hyperspectral band selection was proposed to reduce the redundant components of HSIs, which projected the band vectors into an accurate manifold space. However, since only spectral information was considered in these dimensionality reduction approaches, the classification maps were usually contaminated by the \lq\lq noisy\rq\rq\ behavior of the labeled pixels.\par
To overcome the aforementioned issue, many spectral-spatial classification methods were developed. For example, in \cite{Jon_EMAP}, the morphological profiles (MP) were first applied for classification of HSIs via a series of opening and closing operations with a structuring element of the increasing size. Based on the MP, a large number of HSI classification approaches have been investigated, such as extended morphological attribute profiles (EMAPs) \cite{EMAP_ICA}, parameter-adaptive EMAPs \cite{pa_EMAP}, and random subspace ensemble-based EMAPs \cite{7064745}. In \cite{HONG201935}, the graph learning technique was first utilized to propagate the labels for the task of hyperspectral dimensionality reduction, achieving the state-of-the-art classification performance for remote sensing images. In \cite{IFRF}, an edge-preserving filtering technology was first developed for feature extraction of HSIs, and then, the multi-scale and hierarchical edge-preserving feature extraction method was extended to construct discriminative features of HSIs \cite{PCAEPFs,7906599}. In addition, the multi-view edge-preserving operations were also developed for classification of HSIs \cite{MEPOs}. In \cite{IID}, intrinsic image decomposition (IID) was used to extract the reflectance component of HSIs, which is more robust to shadow and noise compared to other feature extractors. Since the adjacent pixels tend to be from the homogeneous region,  the superpixel-based IID was developed for classification of HSIs \cite{IID_gu}. In \cite{EPs}, an extinction profiles (EPs) was proposed to extract the features of HSIs via the extinction filters with multilevel decomposition. In \cite{8447427}, a new low rank representation approach was designed for HSI classification, in which a new distance metric was used to capture the local similarity of pixels. To increase the robustness of the extracted features, a multi-scale total variation method was utilized to build the structural features of HSIs \cite{MSTV}. These approaches confirmed the fact that feature extraction techniques are effective tools in increasing the classification accuracy.\par
Recently, many deep learning models have been proposed for HSI classification, such as convolutional neural networks (CNNs) \cite{CNNs,CNN_Kang}, recurrent neural network (RNN) \cite{Hang_RNN,RNN2}, and generative adversarial networks \cite{8307247,8661744}. Chen \emph{et al.} used CNN for the first time for deep feature extraction of HSIs via using several convolutional and pooling layers \cite{CNNs}. Hang \emph{et al.} developed a cascaded RNN method using gated recurrent units to reduce the redundant and complementary information of HSIs. Zhu \emph{et al.} applied the generative adversarial network for classification of HSIs, in which the generator (e.g., a conv-deconv network with skip connectors) was used to simulate fake data while the discriminator (a CNN-based classifier) was used to differentiate between fake and real data \cite{8307247}. However, most deep learning-based approaches require massive training data to accurately learn a great deal of parameters. To alleviate this issue, some effective classification frameworks have been developed to classify HSIs, such as semi-supervised classification frameworks \cite{semi_CNN}, active learning frameworks \cite{CNNs2}, and ensemble learning frameworks \cite{8721080}. These classification approaches proved that the spatial information obtained by shallow or deep feature extraction is of vital importance for classification of HSIs. \par
Considering the importance of the spatial information in hyperspectral image classification, in this study, a novel classification framework based on the fusion of dual spatial information is proposed to flexibly integrate class probabilities obtained by the feature extraction and the spatial optimization. The main motivation of this integration is that the feature extraction tends to preserve large-scale objects while the spatial optimization can model the small-scale objects. By fusing the complementarity between dual spatial information, the classification performance can be greatly improved. Specifically, the spectral bands of the original HSI are first decreased with an averaging-based method. Then, the dual spatial information is generated by spatial feature extraction and spatial optimization. In order to extract highly discriminative spatial features, an adaptive texture smoothing method is proposed for extracting the structural profile of HSIs. Then, a support vector machine (SVM) classifier is performed on the extracted structural profile to obtain the class probability of the input. In the spatial optimization, the SVM classifier is performed on the dimension reduced data to obtain the class probability, and the extended random walker (ERW) is utilized to optimize the class probability by using correlation among neighboring pixels. Finally, the class probabilities obtained by exploiting dual spatial information are integrated together by a decision fusion rule. More precisely, the main contributions of this work are listed as follows:

\begin{itemize}
\item{} This work proposes a novel feature extraction method, i.e., structural profile (SP), for HSIs, which is based on adaptive texture smoothing. The SP aims to preserve the geometrical characteristics and smooth out unimportant spatial textures. The discriminative ability belonging to different land covers could be greatly amplified.
\item{} A general hyperspectral classification framework is proposed based on the fusion of dual spatial information, which can effectively improve the classification performance of HSIs. In the proposed framework, the spatial information of the original image can be fully utilized by considering the correlation among adjacent pixels.
\item{} To validate the generalization capability and advantage of the proposed method, experiments are conducted on three different scenes, including a crop scene, a mineral scene, and an urban scene. Experimental results demonstrate that our method consistently outperforms other state-of-the-art methods in all cases. The corresponding code of the proposed method will be made available at the author's Github repository \footnote{https://github.com/PuhongDuan}.
\end{itemize}

The reminder of this work is summarized as follows. Section \ref{sec:proposed} shows the proposed methodology. Section \ref{sec:exp} is on the experiments and analyses. Finally, conclusions are provided in Section \ref{sec:cons}.\par

\section{Proposed Method}
\label{sec:proposed}
\begin{figure*}[!tp]
\centering
\centerline{{\includegraphics[scale=0.43]{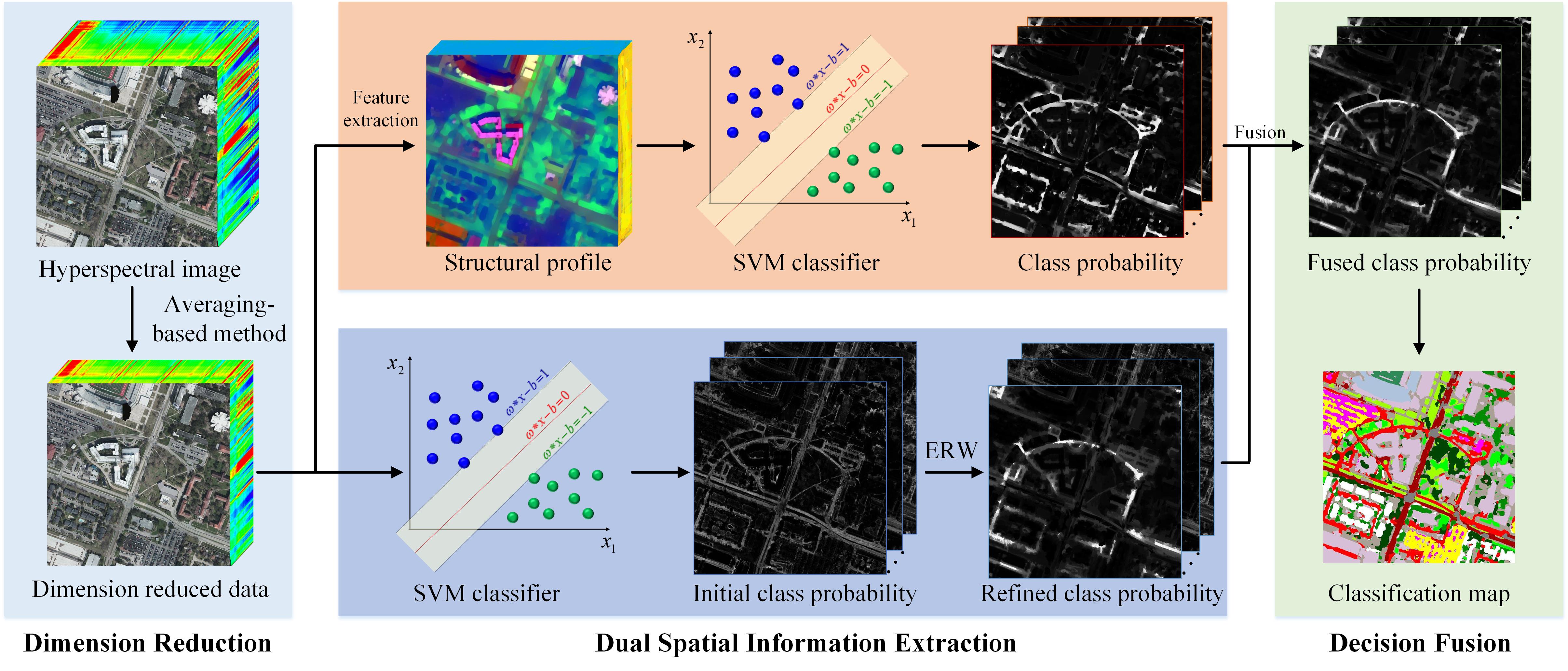}}}
\caption{Flow chart of the proposed classification framework.}
\label{schematic}
\end{figure*}

Fig. \ref{schematic} presents the flow chart of the proposed method, which is comprised of three key steps: Dimension reduction, dual spatial information extraction, and decision fusion. The detailed steps are shown below.
\subsection{Dimension Reduction}
In order to decrease the execution time of the subsequent feature extraction step, the spectral dimension of HSI is reduced with an averaging-based fusion method. Specifically, the original image $\mathbf{O}$ is first divided into $M$ subsets along with spectral dimension. Then, the averaging-based fusion rule is performed on each subset to reduce the dimension of the data $\mathbf{I}$. Although many advanced dimensionality reduction methods \cite{9082155} have been proposed, the averaging-based dimensionality reduction \cite{MEPOs} is found to be an effective tool in terms of computational efficiency and preserving the pixel reflectance.
\subsection{Dual Spatial Information Extraction}
 The spatial feature extraction can well preserve the large-scale objects while easily remove the small-scale objects due to the inaccurate smoothing parameters or low discriminative capability. On the contrary, the spatial probability optimization can highlight the small-scale objects because of considering the relationship of adjacent pixels. Therefore, by jointly merging both levels of class probabilities, the classification accuracy is expected to be increased. In this work, a dual spatial information extraction method is proposed based on pre-processing spatial feature extraction and post-processing spatial probability optimization. In the pre-processing spatial feature extraction stage, the discriminative features of an HSI are constructed by using the proposed SP. In the post-processing spatial optimization stage, the extended random walker (ERW) is considered to model the spatial correlation between neighboring pixels. In more detail, the proposed dual spatial information extraction method is shown as follows:
\subsubsection{Spatial feature extraction} The dimension reduced HSI $\mathbf{I}$ is modeled as a combination of a structural profile $\mathbf{S}$ and texture profile $\mathbf{T}$. The aim of this model is to estimate the structural profile $\mathbf{S}$ using the known $\mathbf{I}$. However, this problem is an ill-posed inverse problem. A common strategy is to introduce image priors, such as low rankness, total variation, and spatial similarity \cite{DUAN2020359,LSM,7926387}. In this work, an adaptive texture smoothing model is proposed:
\begin{equation}
\label{eq:mm}
\arg \mathop {\min }\limits_{\mathbf{S}} \;\left\| {\mathbf{S} - \mathbf{I}} \right\|_2^2 \odot {\omega}  + \lambda {\left\| \mathbf{S} \right\|_{TV}}
\end{equation}
where ${\omega}$ is a weight which controls the similarity of adjacent pixels. $\lambda$ is a free parameter.\par
To smooth out the texture information of HSI, we construct a local polynomial $p = \sum\nolimits_{l = 1}^m {{c_l}{p_l}}$ of degree $L$, denoted as ${\prod _L}$, where $m$ is the number of elements in ${\prod _L}$. For each pixel $\bm{x}$, assume $
\Omega (\bm{x}) = \{ {\bm{x}_1},{\bm{x}_2},...,{\bm{x}_N}\}$ be a finite point set around $\bm{x}$ in $\mathbf{I}$, where $N$ is the number of pixels in $\Omega (\bm{x})$. The structural profile $\mathbf{S}$ is obtained by $\mathbf{S}(\bm{x}):= p(\bm{x})$ for each $\bm{x}\in\Omega$. Then, the high order approximation $\bm{x}$ can be obtained by minimizing the following problem:
\begin{equation}
 \mathop {\arg\min }\limits_{p \in {\prod _L}} \left\{ {\sum\limits_{i = 1}^N {\left\| {p({\bm{x}_i}) - \mathbf{I}({\bm{x}_i})} \right\|_2^2 {\omega} (\bm{x},{\bm{x}_i}) + \lambda {{\left\| {\nabla p({\bm{x}_i})} \right\|}_{TV}}} \;} \right\}
\end{equation}
where ${\prod _L} = \{ {\bm{x}^\alpha }:\bm{x} \in {R^2},\alpha  \in {Z_ + },{\left| \alpha  \right|_1} \le L\}$ indicates the polynomial space with degree $\le L$, ${\omega}$ decides how many pixels $\mathbf{I}(\bm{x}_i)$ devote to the construction of the polynomial approximation $p(\bm{x}_i)$.
\begin{equation}
{\omega} ({\bm{x}_i},\bm{x}) = \exp (\frac{{\sum\limits_{\bm{y} \in Y({\bm{x}_i})} {\left\| {\mathbf{I}({\bm{x}_i} + \bm{y}) - \mathbf{I}(\bm{x} + \bm{y})} \right\|_2^2{G_\sigma }(\left\| \bm{y} \right\|)} }}{{h_0^2}})
\end{equation}
where $G_\sigma $ is the Gaussian function with standard deviation $\sigma$. $h_0$ is the scale parameter that is set as 1, and $Y(\cdot)$ is a small region used to compare the patches around $\bm{x}_i$ and $\bm{x}$. The model (\ref{eq:mm}) can be expressed as:
\begin{equation}
\label{eq:m2}
\mathop {\arg \min }\limits_{p \in {\prod _L}}\left\{ \sum\limits_{i = 1}^N {\left\| {p({\bm{x}_i}) - \mathbf{I}({\bm{x}_i})} \right\|_2^2 {\omega} (\bm{x},{\bm{x}_i})} \; + \lambda {\left\| {\nabla p({\bm{x}_i})} \right\|_1}\right\}.
\end{equation}
The split Bregman iteration algorithm \cite{Sbregman} is used to tackle the problem (\ref{eq:m2}) due to its simplicity, which results in the following updates:\par
\textbf{Step 1}: Update ${p^{k + 1}}({\bm{x}_i})$
\begin{equation}
\label{eq:pp}
\begin{aligned}
{p^{k + 1}}({\bm{x}_i}) = &\arg \mathop {\min }\limits_{p \in {\prod _L}} \sum\limits_{i = 1}^N {\left\| {p({\bm{x}_i}) - \mathbf{I}({\bm{x}_i})} \right\|_2^2 \omega (\bm{x},{\bm{x}_i})} \; +\\
&\lambda \left\| {{d^k}({\bm{x}_i}) - \nabla p({\bm{x}_i}) - {b^k}({\bm{x}_i})} \right\|_2^2 {\omega} (\bm{x},{\bm{x}_i}).
\end{aligned}
\end{equation}
To illustrate the solution of the sub-problem (\ref{eq:pp}) in a simple way, the weight norm is defined as:
\begin{equation}
\left\| \mathbf{g} \right\|_{\omega ,\bm{x}}^2 = \sum\nolimits_{i = 1}^N {\left\| {g({\bm{x}_i})} \right\|} _2^2\omega (\bm{x},{\bm{x}_i}).
\end{equation}
Furthermore, let
\begin{equation}
\mathbf{I}: = \{ \mathbf{I}({\bm{x}_i}):{\bm{x}_i} \in \Omega(\mathbf{x})\}
\end{equation}
\begin{equation}
\mathbf{p}: = \{ p({\bm{x}_i}):{\bm{x}_i} \in \Omega(\mathbf{x})\}.
\end{equation}
Likewise, the same notation is for $\mathbf{b}$ and $\mathbf{d}$. Since we construct $\mathbf{p}$ in ${\prod _L}$ of dimension $m$, $\mathbf{p}$ can be written as a linear combination of the basis $\{ {p_l},l = 1,2,...,m\}$:
\begin{equation}
\mathbf{p}(\bm{x}) = \sum\nolimits_{l = 1}^m {{c_l}{p_l}(\bm{x})}.
\end{equation}
Thus, it can be observed that Eq. (\ref{eq:pp}) is a least squares problem, which can be solved for the coefficients $\bm{c} = \{ {c_l}:l = 1,2,...,m\}$ of the polynomial $p$.
To this end, let us use the energy function with variable $\bm{c}$:
\begin{equation}
\label{eq:pp2}
\begin{aligned}
\varphi (\bm{c}) = &\left\| {\mathbf{p} - \mathbf{I}} \right\|_{\omega ,{\bm{x}}}^2 + \lambda \left\| {{\mathbf{d}_x} - {\nabla _x}\mathbf{p} - {\mathbf{b}_x}} \right\|_{\omega ,{\bm{x}}}^2 +\\
&\lambda \left\| {{\mathbf{d}_y} - {\nabla _y}\mathbf{p} - {\mathbf{b}_y}} \right\|_{\omega ,{\bm{x}}}^2
\end{aligned}
\end{equation}
The coefficients $\bm{c}$ can be obtained by calculating the derivative of the Eq. (\ref{eq:pp2}) and setting it to zero:
\begin{equation}
\label{eq:d}
\frac{{\partial \varphi (\bm{c})}}{{\partial {c_l}}} = 0,\quad l = 1,2,...,m.
\end{equation}
Let $\mathbf{E}(i,l): = {p_l}({\bm{x}_i}),\;{\bm{x}_i} \in \Omega(\mathbf{x}),\;l = 1,2,...,m$ and ${{\mathbf{D}}_\omega }(i,i): = \omega ({{\bm{x}}_i},{\bm{x}}),\quad {{\bm{x}}_i} \in \Omega(\mathbf{x})$.\par
By using these notations, the solution of Eq. (\ref{eq:d}) is given by:
\begin{equation}
\begin{aligned}
\bm{c} = &(\mathbf{I}{\mathbf{D}_\omega }\mathbf{E} + 2\lambda ({\mathbf{d}_x} - {\mathbf{b}_x}){\mathbf{D}_\omega }{\mathbf{E}_x} + 2\lambda ({\mathbf{d}_y} - {\mathbf{b}_y}){\mathbf{D}_\omega }{\mathbf{E}_y})\\&{({\mathbf{E}^T}{\mathbf{D}_\omega }\mathbf{E} + 2\lambda \mathbf{E}_x^T{\mathbf{D}_\omega }{\mathbf{E}_x} + 2\lambda \mathbf{E}_y^T{\mathbf{D}_\omega }{\mathbf{E}_y})^{ - 1}}.
\end{aligned}
\end{equation}
Accordingly, the solution $p^{k+1}$ is obtained by:
\begin{equation}
{p^{k + 1}}: = \sum\limits_{l = 1}^m {{c_l}{p_l}}.
\end{equation}\par
\textbf{Step 2}: Update ${d^{k + 1}}({\bm{x}_i})$\par
\begin{equation}
\mathop {\arg\min }\limits_d {\left| {d({\bm{x}_i})} \right|_1} + \lambda \left\| {d({\bm{x}_i}) - \nabla {p^{k + 1}}(\bm{x}) - {b^k}(\bm{x})} \right\|_2^2
\end{equation}
This $L_1$-norm-induced problem can be solved effectively via the soft-thresholding method as:
\begin{equation}
{d^{k + 1}}({\bm{x}_i}){ = }{{soft}}(\nabla {p^{k + 1}}({\bm{x}_i}) + {b^k}({\bm{x}_i}),{1 \mathord{\left/
 {\vphantom {1 \lambda }} \right.\kern-\nulldelimiterspace} \lambda })
 \end{equation}
 where ${{soft}}(a,b)=sign(a)\cdot max(\left|a\right|-b,0)$.\par
\begin{figure}[!tp]
\centering
\centerline{{\includegraphics[scale=0.5]{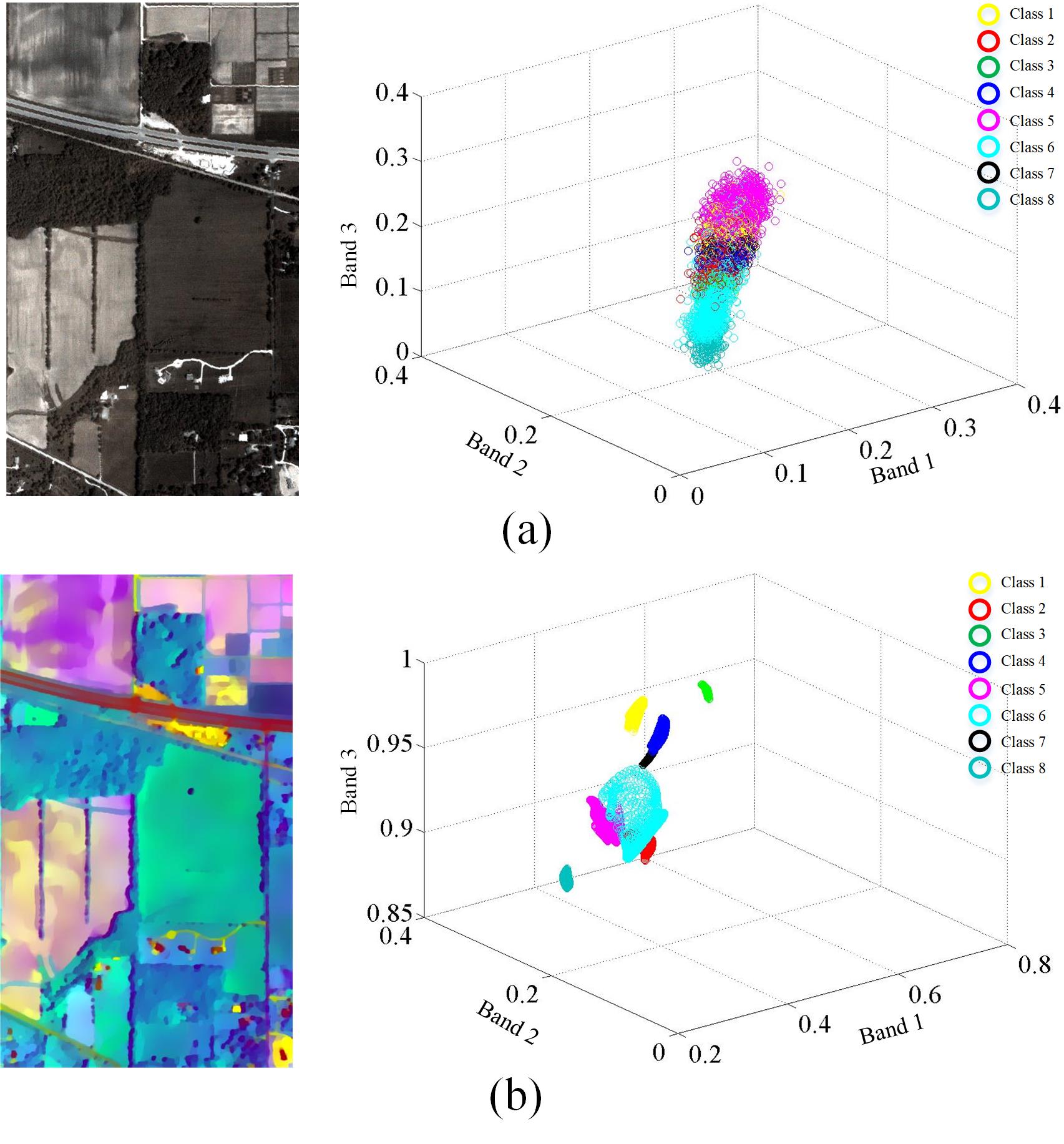}}}
\caption{This example illustrates the between-class distances on training samples (for an intuitive display, only the
first eight classes are given.). (a) The first three bands of an original HSI and the discrimination of different classes. (b) The first three bands of the extracted structural profile and the discrimination of different classes.}
\label{dis}
\end{figure}

 \textbf{Step 3}: Update ${b^{k + 1}}({\bm{x}_i})$
 \begin{equation}
{b^{k + 1}}({\bm{x}_i}){= }{b^k}({\bm{x}_i}) + \nabla {p^{k + 1}}({\bm{x}_i}) - {d^{k + 1}}({\bm{x}_i})
\end{equation}
The above three steps are iterated until convergence obtained on each pixel. Accordingly, the initial structural profile can be obtained. In order to reduce the spectral redundancy and further decrease discrimination between different objects in HSI, the kernel principal component analysis (KPCA) \cite{KPCA} is applied on the initial SP to obtain more discriminative structural profile $\mathbf{S}$, and $K$ principal components are preserved for classification. The aforementioned steps aim to yield the SP from the original HSI, serving as pre-processing spatial feature extraction. As shown in Fig. \ref{dis}, the discrimination belonging to different objects can be greatly improved. \par
Then, the SVM classifier is conducted on the extracted SP to obtain the class probability ${\mathbf{C}}_1$.
\subsubsection{Spatial probability optimization}
Although the feature extraction technique can well model the spatial information of HSIs, the optimal parameter is hard to be determined in real applications. In this situation, some small objects are easily removed in the feature extraction stage. To alleviate this issue, the post-processing spatial probability optimization is proposed, which is an effective complementary way to avoid the over-smoothing phenomenon caused by the feature extraction. More precisely, the dimension reduced data $\mathbf{I}$ is first fed into the spectral classifier, i.e., SVM, to obtain the initial class probability $\hat{\mathbf{C}}_2$. Then, the ERW \cite{ERW} is adopted to optimize the initial probability by modeling the spatial patterns among adjacent pixels. Specifically, the optimized probability of different classes can be achieved by solving the following energy function:
\begin{equation}
\label{eq:erw}
{E^t}({\mathbf{Q}_t}) = E_{spatial}^t({\mathbf{Q}_t}) + \gamma E_{aspatial}^t({\mathbf{Q}_t})
\end{equation}
where $t$ is different classes. Eq. (\ref{eq:erw}) is composed of the spatial and the aspatial terms. The spatial term $\mathbf{E}_{spatial}^t$ is used to model the spatial dependence among adjacent pixels.
\begin{equation}
E_{spatial}^t({\mathbf{P}_t}) = \mathbf{P}_t^T\mathbf{L}{\mathbf{P}_t}
\end{equation}
where $\mathbf{L}$ stands for the Laplacian matrix of a weighted graph.
\begin{equation}
{\mathbf{L}_{ij}} = \left\{ \begin{array}{l}
\sum {{e^{ - \beta {{({v_i} - {v_j})}^2}}}} \quad \;if\;i = j\\
 - {e^{ - \beta {{({v_i} - {v_j})}^2}}}\quad if\;i\;and\;j\;are\;adjacient\;pixels\\
0\quad \quad \quad \quad \quad \quad \;otherwise
\end{array} \right.
\end{equation}
where the first component of original HSI obtained by KPCA is used to construct the weighted graph. $v_i$ denotes the $i$th pixel of the first component.
The second term $E_{aspatial}^t$ integrates the initial probability $\hat{\mathbf{C}}_2$.
\begin{equation}
E_{aspatial}^t({\mathbf{Q}_t}) = \sum\limits_{q = 1,q \ne t} {\mathbf{Q}_q^T{\mathbf{\Lambda} _q}{\mathbf{Q}_q} + {{({\mathbf{Q}_t} - 1)}^T}{\mathbf{\Lambda} _t}({\mathbf{Q}_t} - 1)}
\end{equation}
where $\mathbf{\Lambda}_t$ is a diagonal matrix, where diagonal values consist of the initial probability $\hat{\mathbf{C}}_2$. Finally, the refined probability $\mathbf{C}_2$ is obtained by choosing the maximum of $\mathbf{Q}_t$.

\subsection{Decision Fusion}
When the class probabilities of the two stages are obtained, the weighted decision fusion rule is considered to merge the class probabilities so as to obtain the final class label. In this way, the class probability belonging to the same object is higher than ones of other land covers, and thus, different objects can be more accurately identified.
In more detail, the final label for each pixel is decided based on the maximum probability.
 \begin{equation}
\mathbf{C}=\mathop {argmax}\limits_i \{ \mu \mathbf{C}_1^i + (1 - \mu )\mathbf{C}_2^i\}
\end{equation}
where $\mathbf{C}^{i}_{1}$ is the class probability in the spatial feature extraction stage. $\mathbf{C}^{i}_{2}$ denotes the class probability in the spatial probability optimization stage. $\mathbf{C}$ is the final classification result. $\mu$ is a free parameter.

\section{Experiments}
\label{sec:exp}
To verify the effectiveness of the proposed approach, three hyperspectral data sets have been used. The data sets have been selected from different hyperspectral applications (i.e., crop area mapping, mineral mapping, and urban mapping) to assess the generalization capability of our method. In the following section, the used data sets are first described. Then, the experimental setup is given. Next, the discussion about the parameters and different components is given. Finally, the experimental results are described.
\subsection{Data sets}
To validate the performance of our method, three representative hyperspectral images from different imaging scenes are employed, which are briefly described below:\par
\begin{figure}[!tp]
\centering
\centerline{\subfigure[]{\includegraphics[scale=0.1]{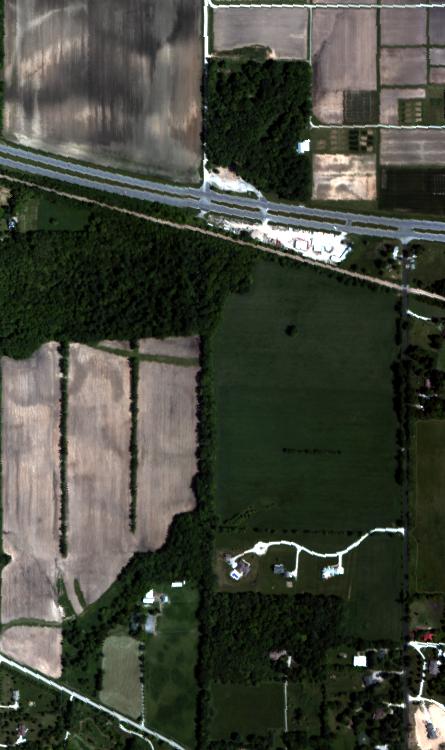}}
			\subfigure[]{\includegraphics[scale=0.1]{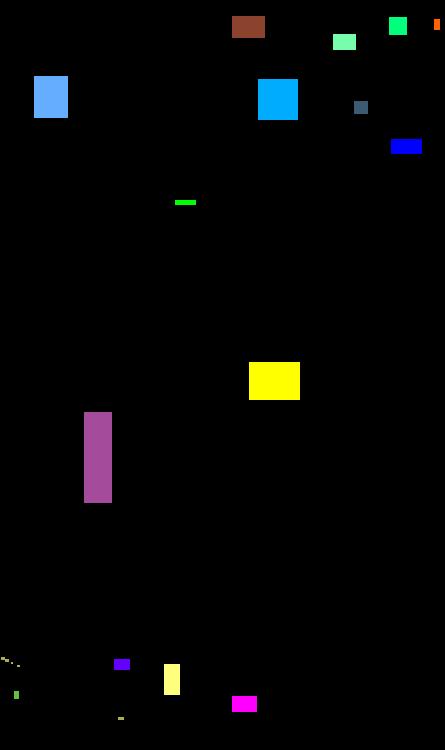}}
            \subfigure[]{\includegraphics[scale=0.1]{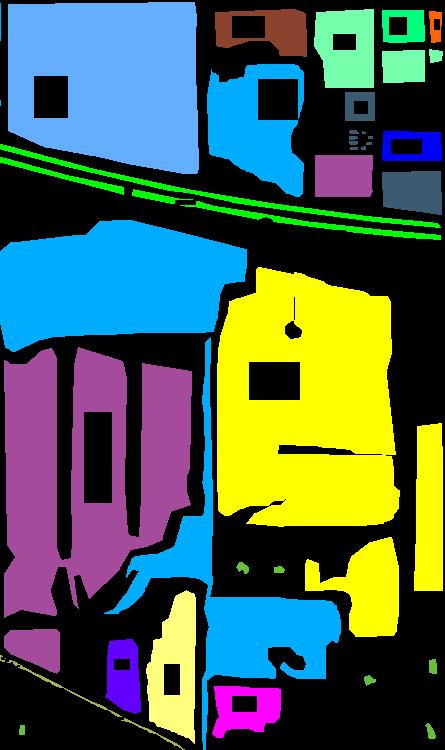}}
           \subfigure[]{\includegraphics[scale=0.25]{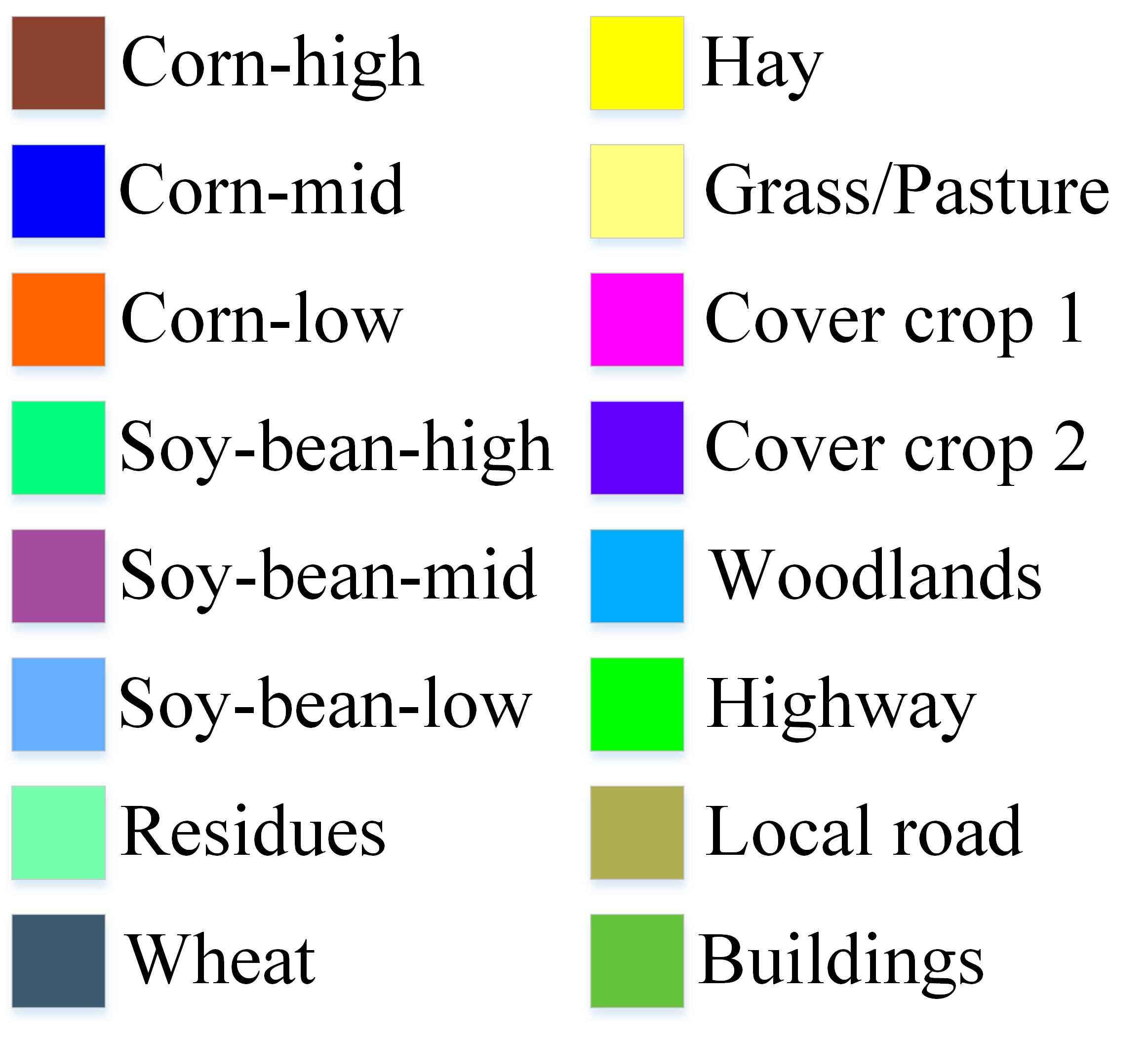}}}
\caption{Crop scene: Indian Pines 2010 data set. (a) Color composite. (b) Training samples. (c) Testing samples. (d) Labels.}
\label{India}
\end{figure}

1) Indian Pines 2010 (Crop scene): The Indian Pines 2010 is considered here for crop identification because it contains multiple different crops, such as corn, soy bean and wheat. This image was captured by the ProSpecTIR sensor around Purdue University, Indiana, on 24-25th of May 2010. It consists of 360 spectral channels ranging from 0.4-2.45 $\mu$m, and the spatial size is 445$\times$750 with 2 m spatial resolution. This scene is composed of 16 different land covers, in which most classes are the crops. Fig. \ref{India} shows the RGB composite, training and test samples. The number of training and test samples is shown in Table \ref{tab:India2010}.\par
\begin{figure}[!tp]
\centering
\centerline{\subfigure[]{\includegraphics[scale=0.18]{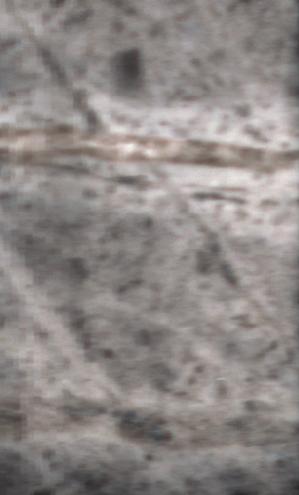}}
			\subfigure[]{\includegraphics[scale=0.18]{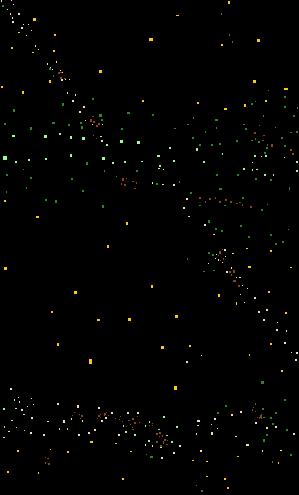}}
            \subfigure[]{\includegraphics[scale=0.18]{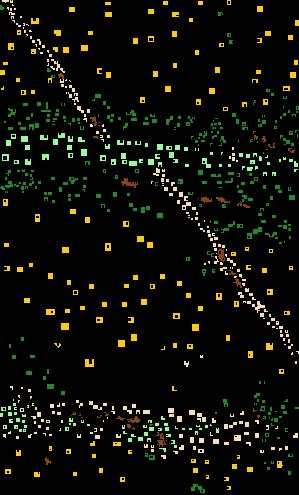}}
           \subfigure[]{\includegraphics[scale=0.24]{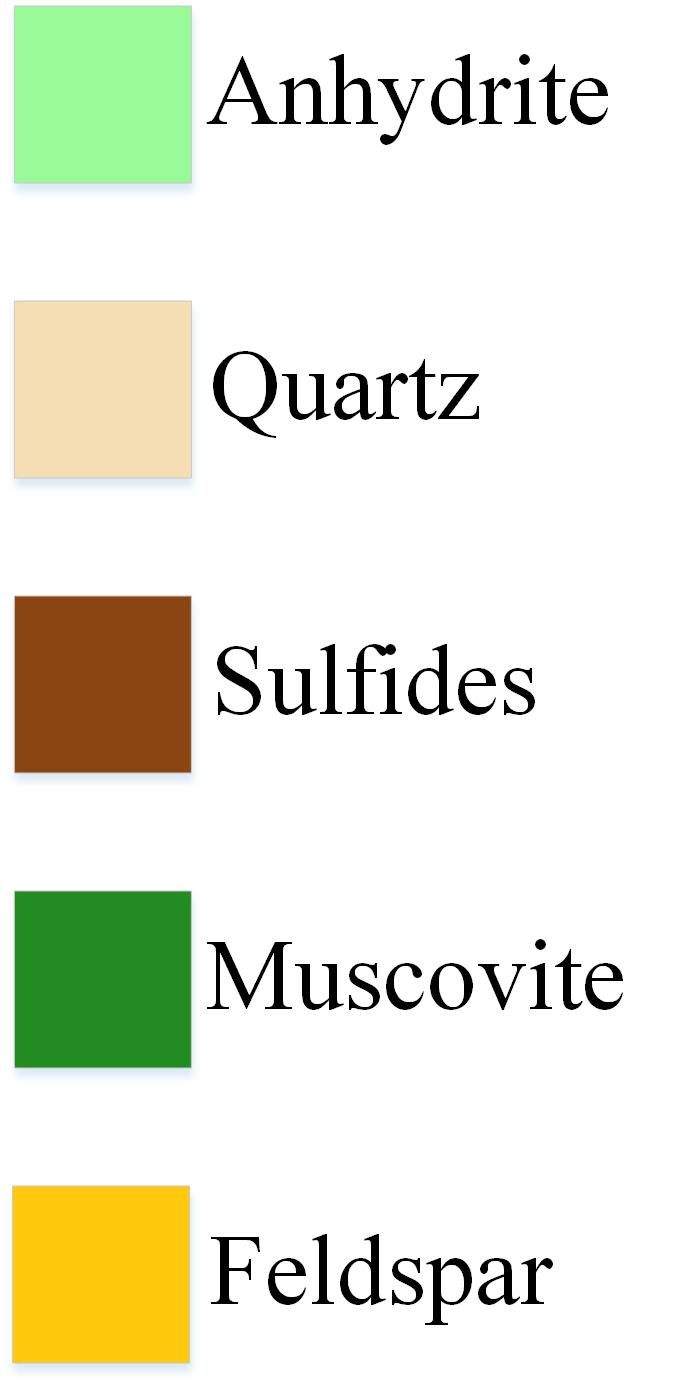}}}
\caption{Mineral scene: TS4-1900 data set. (a) Color composite. (b) Training samples. (c) Testing samples. (d) Labels.}
\label{TS4}
\end{figure}

2) TS4-1900 (Mineral scene): TS4-1900 is a drill-core mineral hyperspectral image, which consists of five different minerals. This image was acquired by the Specim sCMOS sensor in the Lab. The specific imaging platform and environment can be found in \cite{s19122787}. It is also comprised of 480 spectral bands with the wavelength ranging from 0.4 to 1 $\mu$m. The spatial size is 495$\times$299 with spatial resolution of 0.08 mm. Fig. \ref{TS4} depicts the three band color composite, and the corresponding training and test samples. The ground truth used in the experiment is obtained by utilizing Scanning Electron Microscopy-based image analyses using a Mineral Liberation Analyser (SEM-MLA). The number of training and test samples is presented in Table \ref{tab:TS1900}.\par
\begin{figure}[!tp]
\centering
\centerline{{\includegraphics[scale=0.23]{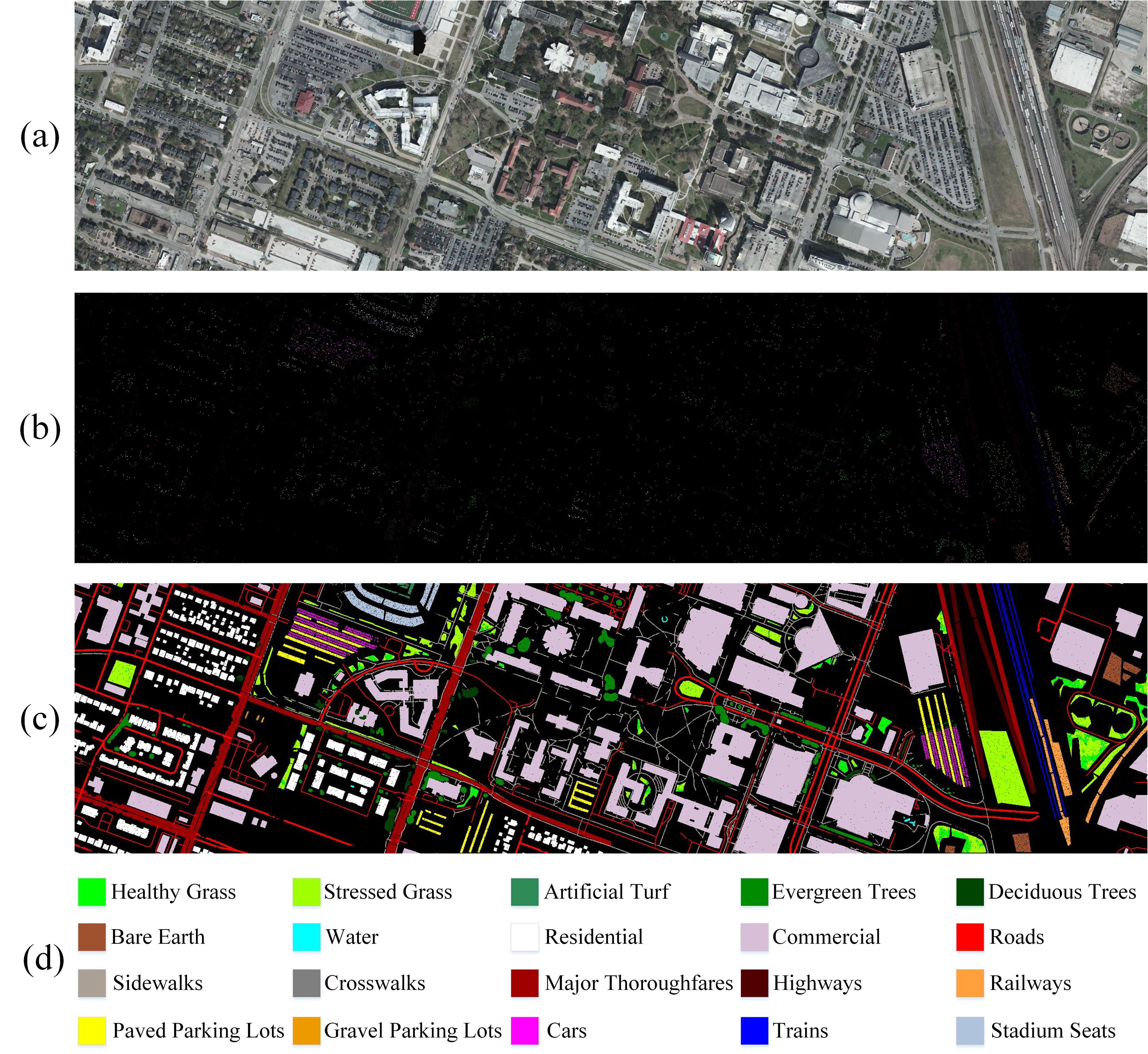}}}
\caption{Urban scene: Houston 2018 data set. (a) RGB image. (b) Training samples. (c) Testing samples. (d) Labels.}
\label{Hu2018}
\end{figure}

3) Houston 2018 (Urban scene): The Houston 2018 is a representative urban scene containing twenty classes of interest, which is released by the IEEE GRSS 2018 data fusion contest \footnote{http://www.grss-ieee.org/community/technical-committees/data-fusion/2018-ieee-grss-data-fusion-contest/}. In this work, only the training portion of the data set distributed for the GRSS data fusion contest is used. This image was acquired by the hyperspectral imager CASI 1500 at the University of Houston on Feb. 16, 2017. It consists of 48 spectral channels ranging from 0.38-1.05 $\mu$m. The spatial size is 601$\times$2384 pixels with the ground sampling distance of 1 m. Fig. \ref{Hu2018} exhibits the RGB image and the sample distribution of the training and test samples. The number of training and test samples is listed in Table \ref{tab:hu18}.

\subsection{Experimental Setup}
1) Evaluation metrics: To quantitatively assess the classification performance of all studied approaches, three popular and extensively used metrics, i.e., overall accuracy (OA), average accuracy (AA), and Kappa coefficient, are employed. OA measures the percentage of total correctly identified pixels. AA is the mean of the percentage of the correctly identified pixels for each land cover. Kappa coefficient calculates the percentage of identified pixels corrected by the number of agreements that would be expected purely by chance. \par
\begin{figure*}[!tp]
\centering
\centerline{{\includegraphics[scale=0.18]{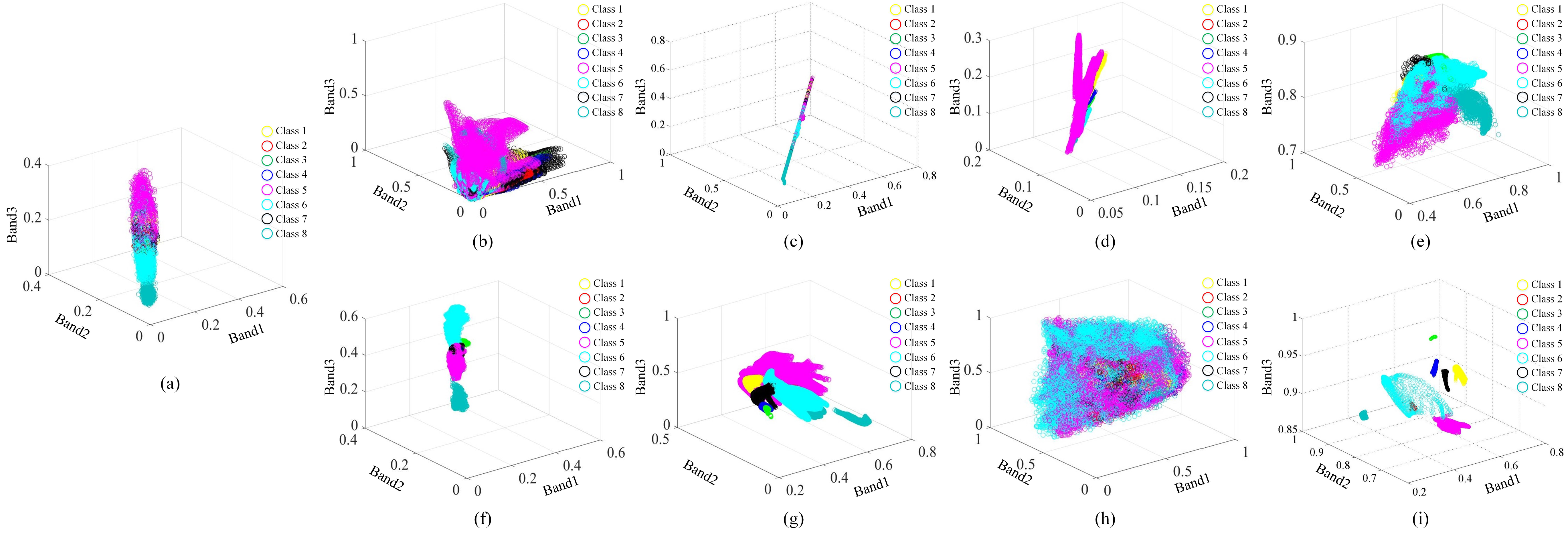}}}
\caption{The scatter diagrams of different methods on the test samples. (a) The original HSI. (b) Gabor \cite{Gabor}. (c) EMP \cite{Jon_EMAP}. (d) IFRF \cite{IFRF}. (e) LRR \cite{OTVCA}. (f) IID \cite{IID}. (g) PCA-EPFs \cite{PCAEPFs}. (h) IAPs \cite{IAPs}. (i) SP.}
\label{separa}
\end{figure*}
2) Competitive approaches: In this work, seven state-of-the-art classification approaches are adopted for comparison, including:
\begin{itemize}
\item{} {\textbf{Support vector machine (SVM)}} \cite{SVM} is performed using the LIBSVM library, where the Gaussian kernel is adopted. The parameters in the SVM are selected using fivefold cross-validation. The kernel width varies between $2^{-5}\sim2^{5}$, and the penalty factor ranges between $10^{-2}\sim10^{4}$.
\item{} {\textbf{Multiple feature learning (MFL)}} \cite{MFL} combines multiple features derived from both linear and nonlinear transformations in a flexible way, in which the morphological profiles and original spectral features are considered in the fusion framework.
\item{} {\textbf{Weighted Markov random field (WMRF)}} \cite{WMRF} aims to characterize spatial information in the hidden field. A total variation regularization model is introduced to smooth out the posterior distribution.
\item{} {\textbf{Superpixel-based classification via multiple kernels (SCMK)}} \cite{SCMK} integrates three kernel features obtained from spectral and spatial aspects. The relationships within and among superpixels are used to construct the discriminative features.
\item{} {\textbf{Orthogonal total variation component analysis (OTVCA)}} \cite{OTVCA} is a low rank representation-based feature extraction method, where the high-dimensional HSI is projected into a low dimension space.
\item{} {\textbf{Random patches network (RPNet)}}  \cite{XU2018344} aims to learn deep convolutional features followed by a spectral classifier, in which the random patches is used as the convolutional kernels.
\item{} {\textbf{Generalized tensor regression (GTR)}} \cite{GTR} is an extended tensorial version of the ridge regression for multivariate labels. The nonnegative prior is used to boost the discrimination of different modes.
\end{itemize}

\begin{figure}[!tp]
\centering
\centerline{\subfigure[]{\includegraphics[scale=0.2]{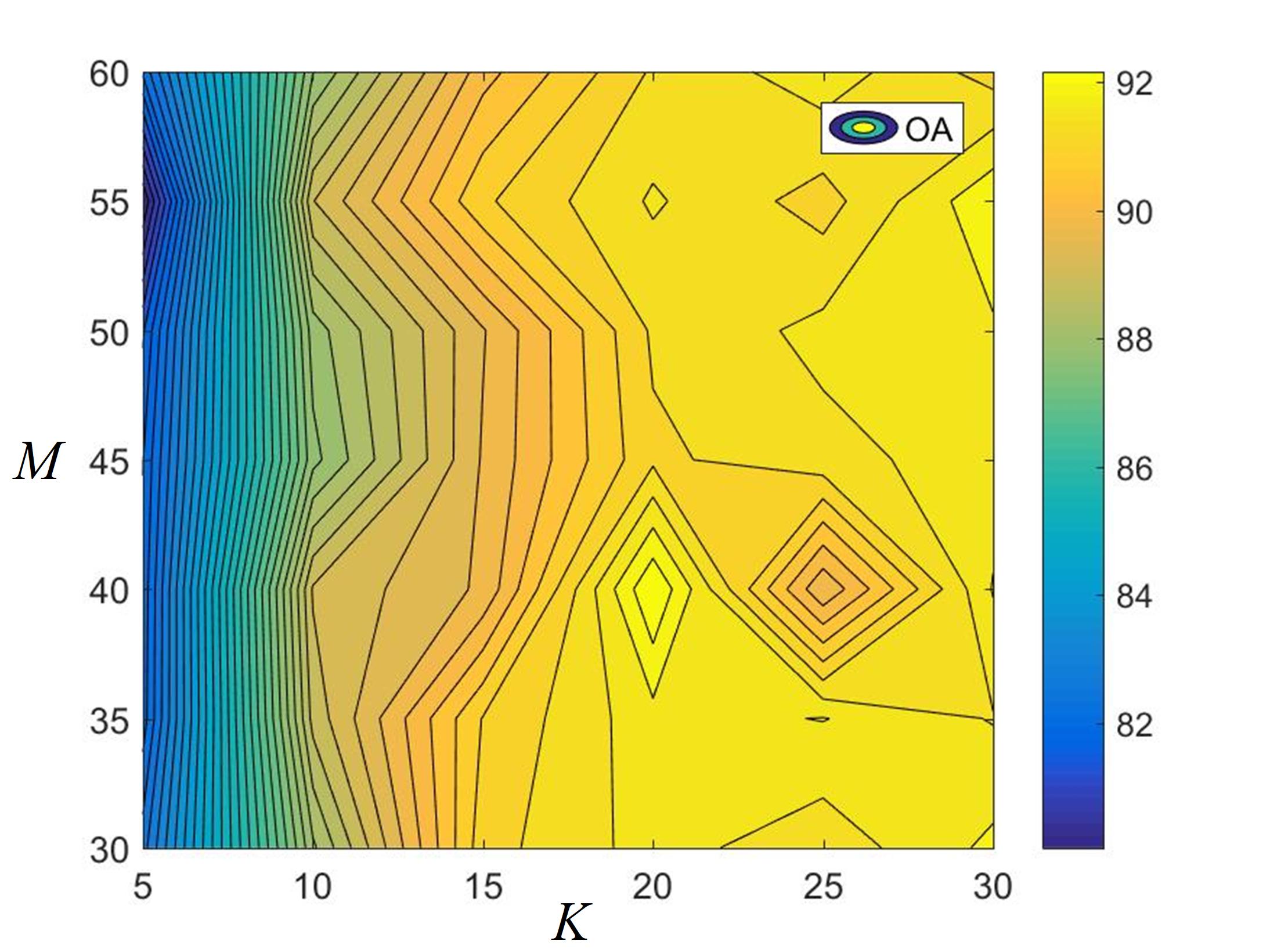}}
			\subfigure[]{\includegraphics[scale=0.2]{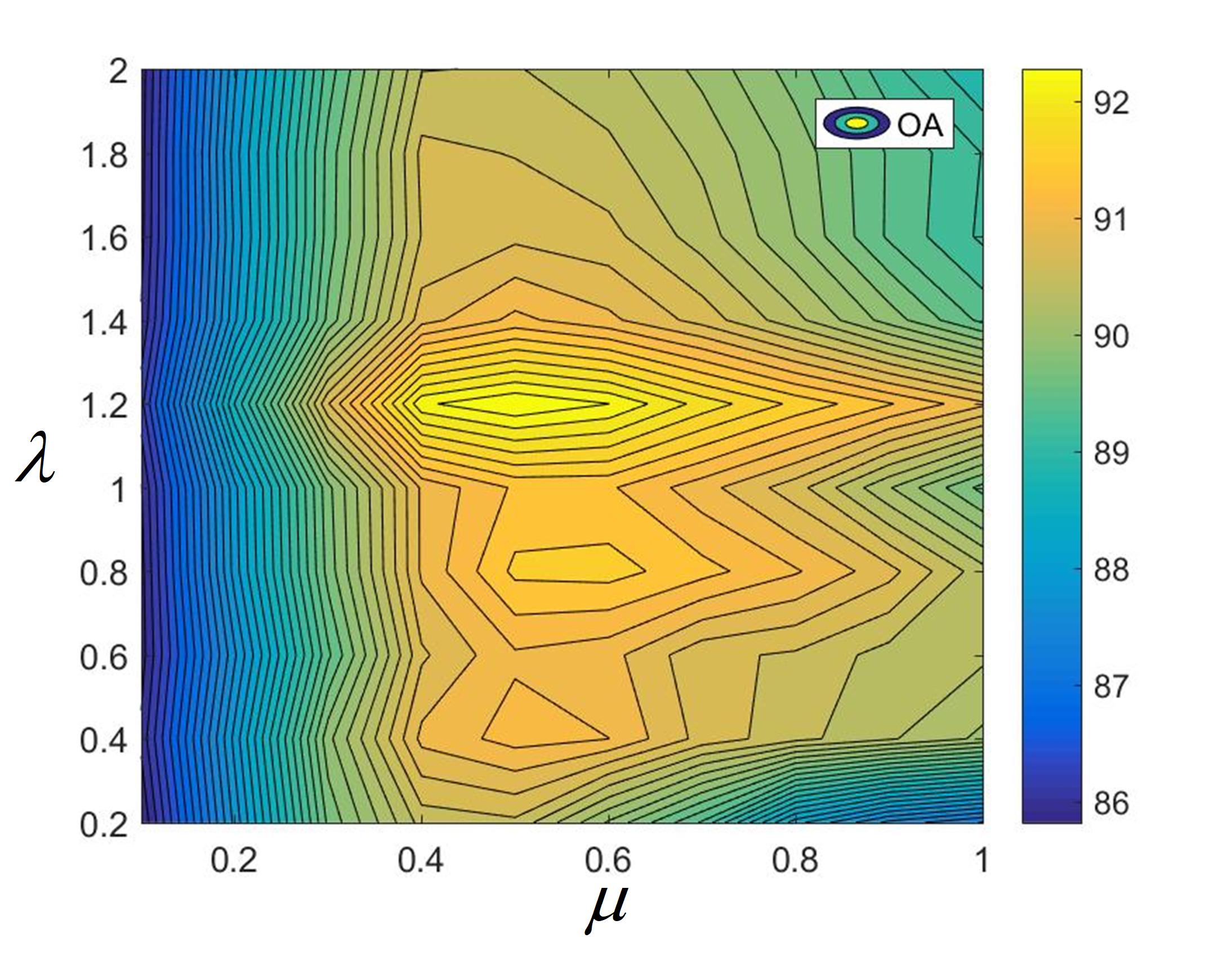}}}
\caption{The influence of the parameters, i.e, $M$, $K$, $\mu$, and $\lambda$, to the classification performance of the proposed method.}
\label{pa}
\end{figure}
3) Parameter analysis: In the proposed method, there are four parameters need to be fixed, i.e., the number of the dimension reduced data $M$, the number of kernel principle components used for classification $K$, the smoothing parameter $\lambda$, and the fusion weight $\mu$. An experiment is performed on the TS4-1900 data set to initialize the aforementioned parameters. When $M$ and $K$ are analyzed, $\lambda$ and $\mu$ are set to be 1 and 0.5, respectively. Fig. \ref{pa} (a) shows the influence of different parameters. It can be observed that when the preserved components $K$ are relatively small, the classification accuracy of our method tends to decrease. Moreover, when $K$ and $M$ are set as 20 and 40, respectively, the proposed method obtains satisfactory performance. Besides, Fig. \ref{pa} (b) shows the influence of parameters $\lambda$ and $\mu$. It is obvious that our method obtains the highest OA when $\mu=0.5$ and $\lambda=1.2$. The following experiments with the default parameter setting verify that the proposed method can achieve satisfactory classification performance on all the data sets.

\subsection{Discussion}
\begin{table}[!tp]
  \centering
  \caption{Classification Performance of Different Feature Extraction Methods on Indian Pines 2010.}
    \begin{tabular}{cccc}\toprule
    Methods & OA    & AA    & Kappa \\\midrule
    Gabor \cite{Gabor} & 55.84 & 69.71 & 43.96 \\
    EMP \cite{Jon_EMAP}   & 74.45 & 71.22 & 69.32 \\
    IFRF \cite{IFRF}  & 56.66 & 72.42 & 45.37 \\
    LRR \cite{OTVCA}  & 89.80  & 78.53 & 87.54 \\
    IID \cite{IID}  & 78.47 & 79.22 & 74.29 \\
    PCA-EPFs \cite{PCAEPFs} & 79.01 & 75.17 & 74.99 \\
    IAPs \cite{IAPs} & 60.01 & 39.88 & 50.89 \\
    SP    & \textbf{91.84} & \textbf{85.68} & \textbf{90.01} \\\bottomrule
    \end{tabular}%
  \label{tab:FE}%
\end{table}%


1) Discriminative feature extraction:
To demonstrate the discriminative ability and effectiveness of the proposed structural profile, several extensively used feature extraction approaches in the hyperspectral community are adopted for comparison, including Gabor filtering (Gabor) \cite{Gabor}, extended morphological profiles (EMP) \cite{Jon_EMAP}, image fusion and recursive filtering (IFRF) \cite{IFRF}, low rank representation (LRR) \cite{OTVCA}, intrinsic image decomposition (IID) \cite{IID}, PCA-based edge-preserving filters (PCA-EPFs) \cite{PCAEPFs}, and invariant attribute profiles (IAPs) \cite{IAPs}. For a fair comparison, except Gabor and EMP feature extractors, the preserved features dimension of other methods is set to be 20. Moreover, all experiments are conducted on the Indian Pines 2010 image with the selected training and test samples (as shown in Fig. \ref{India}), and the SVM classifier is adopted to evaluate the classification performance. \par

Table \ref{tab:FE} shows the accuracies obtained by different methods. As shown in this figure, it can be observed that the proposed SP feature extractor achieves the highest accuracies in terms of the OA, AA, and Kappa among all studied approaches, which demonstrates that the SP is indeed more effective for hyperspectral image classification. The main reason is that the proposed feature extraction method can greatly increase the pixel separability belonging to different objects.\par
 Fig. \ref{separa} shows the scatter diagrams of different methods on the test samples. Based on the separability analysis, several interesting observations can be observed. First, the low discrimination belonging to different classes will result in unsatisfactory classification performance. Taking Fig.  \ref{separa} (b) and (h) as examples, it can be seen that the test samples of different classes are fully intersected with each other, and thus, the classification accuracies of the two methods are relatively low (see Table \ref{tab:FE}). Second, it is found that the SP can effectively reduce the inner-class variation and enhance the inter-class discrepancy compared to other feature extraction approaches. The main reason is that the SP removes unimportant texture components and preserves significant
spatial structures. We can observe from Fig. \ref{separa}(h) that the pixels belonging to the same class tend to be clustered together while the pixels belonging to different classes are separated in different locations. Third, a single-texture-feature extractor for classification of hyperspectral image cannot obtain satisfactory performance such as Gabor and IAPs, and thus, these feature extraction methods are genrally considered as a pre-processing step in most publications.

\begin{figure}[!tp]
\centering
\centerline{{\includegraphics[scale=0.3]{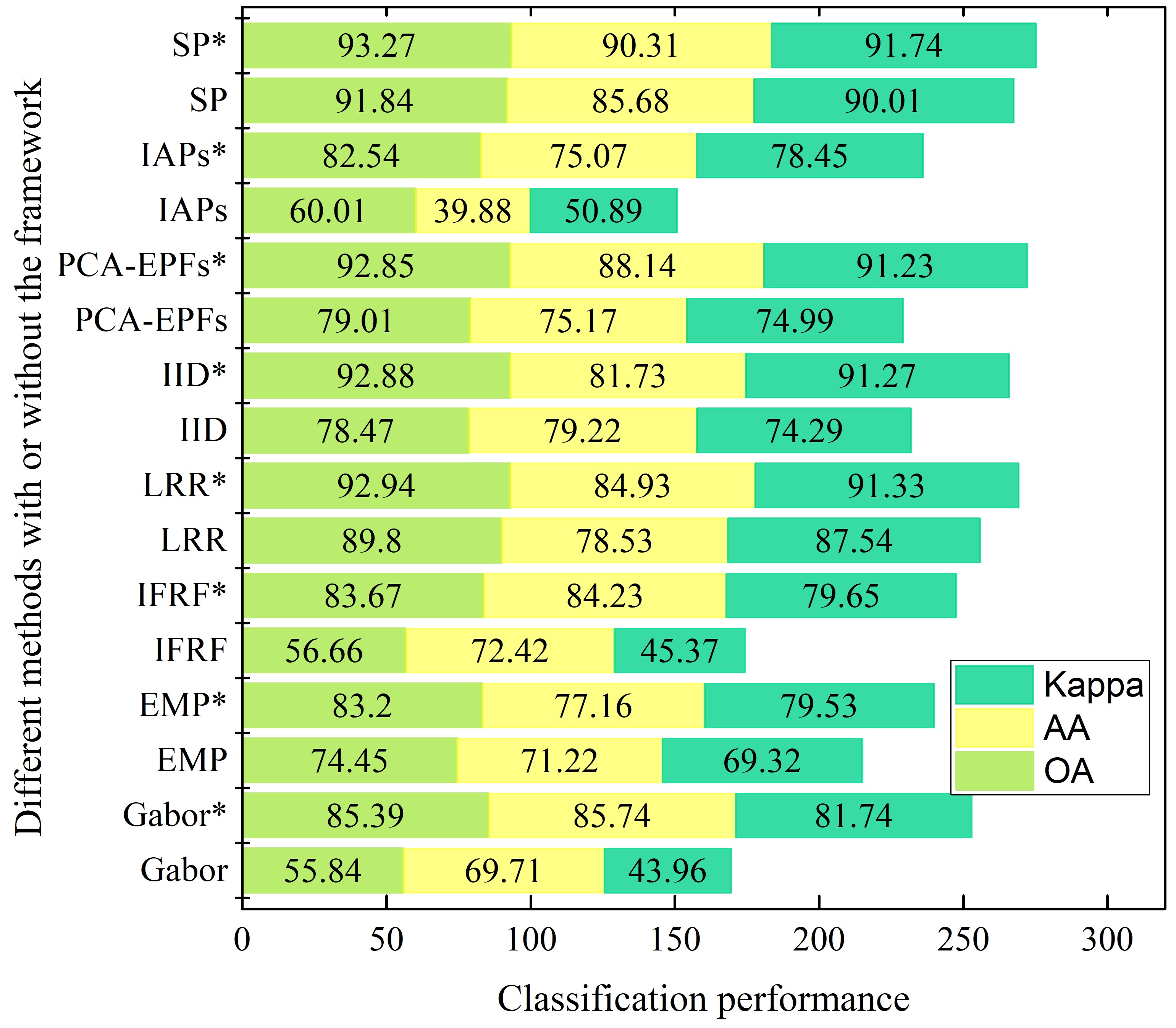}}}
\caption{The classification performance of different feature extraction methods with or without the proposed framework. Notation \lq\lq * \rq\rq\ denotes the feature extraction method with the proposed framework.}
\label{fean}
\end{figure}
2) Classification framework:
In order to illustrate the effectiveness of the proposed classification framework, an experiment is conducted on the Indian Pines 2010 image with or without the proposed framework, in which different feature extractors are used as the pre-processing step for spatial feature extraction. Fig. \ref{fean} shows the classification performance of different feature extraction methods with or without the framework proposed in this work. It can be seen from Fig. \ref{fean} that when the feature extraction technique cannot provide good classification performance due to the low discriminative ability or over-smoothed phenomenon, the proposed framework can greatly improve the classification accuracy (see Gabor and Gabor*). In addition, it is found that when the proposed framework can always produce the better classification performance when the feature extraction method and the spatial probability optimization are combined together. This demonstrates that the dual spatial information-based fusion framework can effectively improve the classification accuracies. The main reason is that the pre-processing feature extraction and post-processing spatial optimization provide complementary characteristics, which can be fully integrated with the proposed framework.

\begin{figure}[!tp]
\centering
\centerline{\subfigure[OA=88.04\%]{\includegraphics[scale=0.12]{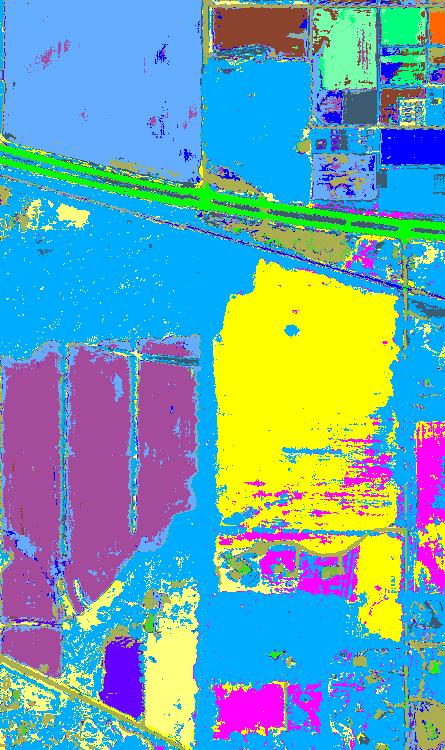}}
			\subfigure[OA=92.10\%]{\includegraphics[scale=0.12]{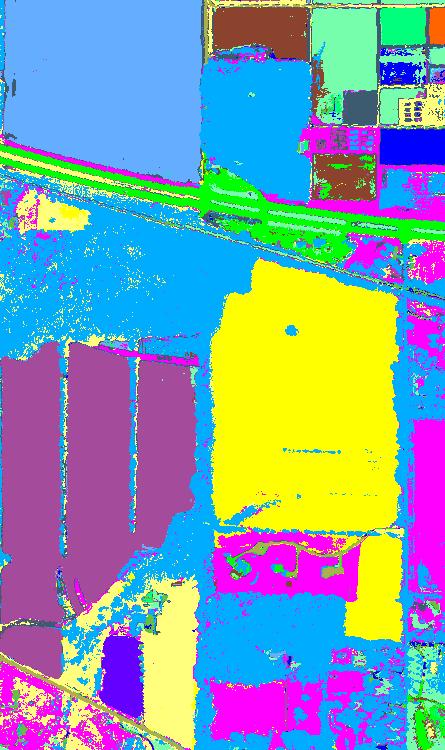}}
            \subfigure[OA=81.07\%]{\includegraphics[scale=0.12]{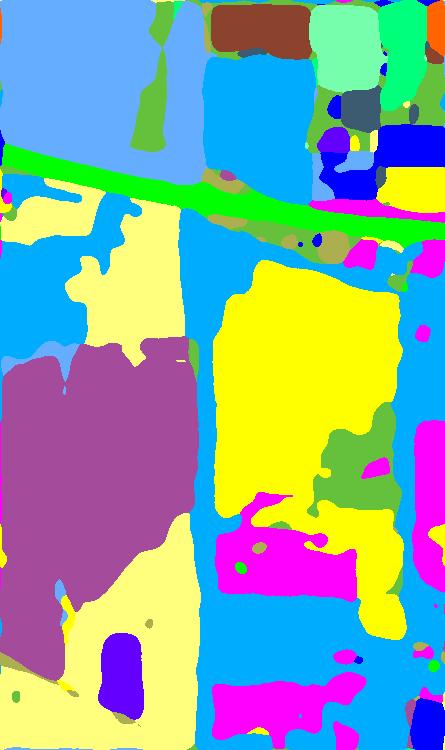}}
            \subfigure[OA=67.61\%]{\includegraphics[scale=0.12]{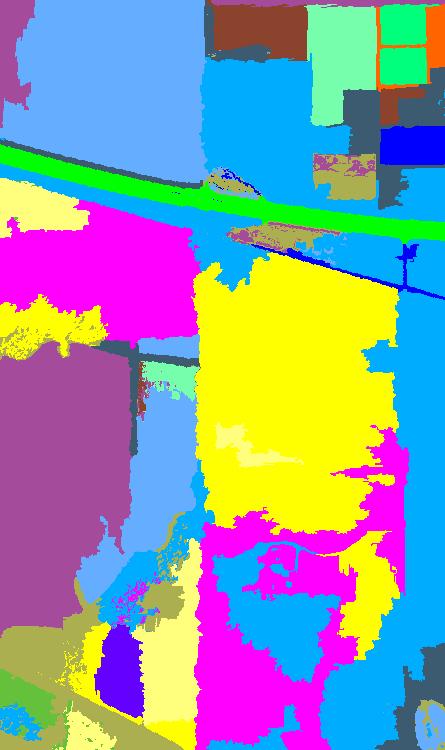}}}
\centerline{\subfigure[OA=87.90\%]{\includegraphics[scale=0.12]{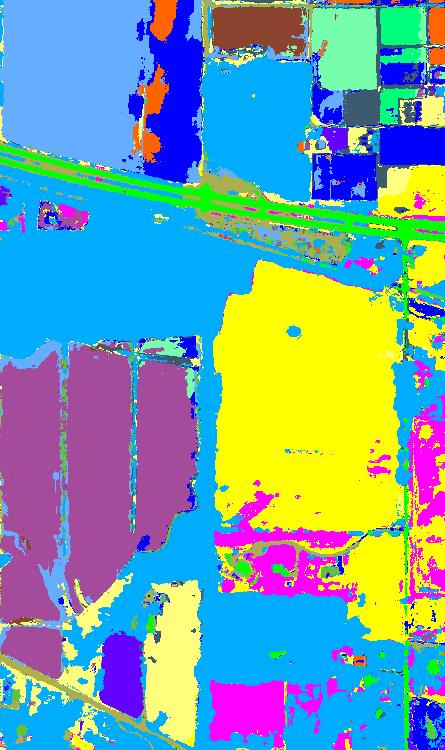}}
			\subfigure[OA=78.26\%]{\includegraphics[scale=0.12]{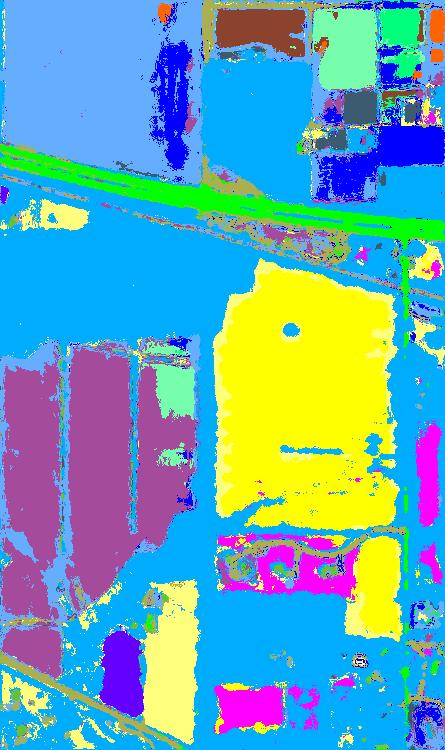}}
            \subfigure[OA=88.89\%]{\includegraphics[scale=0.12]{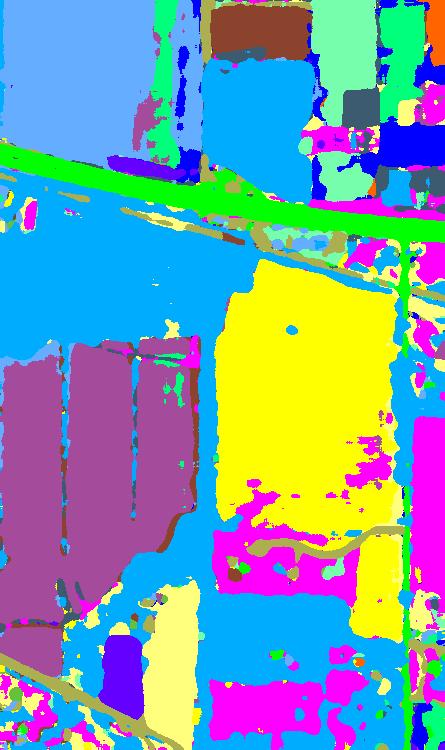}}
			\subfigure[OA=93.27\%]{\includegraphics[scale=0.12]{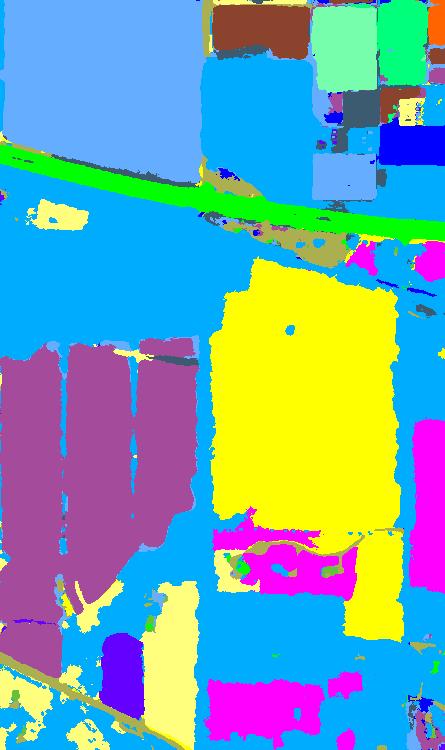}}}
\caption{Classification maps of all studied approaches on the Indian Pines 2010 data set. (a) SVM \cite{SVM}, (b) MFL \cite{MFL}, (c) WMRF \cite{WMRF}, (d) SCMK \cite{SCMK}, (e) OTVCA \cite{OTVCA}, (f) RPNet \cite{XU2018344}, (g) GTR \cite{GTR}, and (h) Our method. }
\label{Indiaresult}
\end{figure}

\begin{table*}[!htbp]
  \centering
  \caption{Classification Performance of All Studied Approaches for Indian Pines 2010 Data Set, Including SVM \cite{SVM}, MFL \cite{MFL}, WMRF \cite{WMRF}, SCMK \cite{SCMK}, OTVCA \cite{OTVCA}, RPNet \cite{XU2018344}, GTR \cite{GTR} and Our Method.}
    \begin{tabular}{ccccccccccc}\toprule
    \multirow{2}[0]{*}[-2pt]{Class name} & \multirow{2}[0]{*}[-2pt]{Training set} & \multirow{2}[0]{*}[-2pt]{Testing set} & \multicolumn{8}{c}{Classification accuracies of different approaches (in \%)} \\\cmidrule(lr){4-11}
          &       &       & SVM   & MFL   & WMRF  & SCMK  & OTVCA & RPNet  & GTR   & Our method \\\midrule
    \multicolumn{1}{>{\columncolor{Corn-high}}c}{Corn-high} & 726   & 2661  & 72.80  & 99.40  & \textbf{100.00} & 97.19  & 97.14  & 83.54  & \textbf{100.00} & 94.66  \\
    \multicolumn{1}{>{\columncolor{Corn-mid}}c}{Corn-mid} & 465   & 1275  & 54.17  & \textbf{100.00} & \textbf{100.00} & \textbf{100.00} & 11.80  & 98.51 & \textbf{100.00} & 99.07  \\
    \multicolumn{1}{>{\columncolor{Corn-low}}c}{Corn-low} & 66    & 290   & 95.97  & 97.24  & \textbf{100.00} & 65.02  & 10.78  & 94.48 & \textbf{100.00} & \textbf{100.00} \\
    \multicolumn{1}{>{\columncolor{Soy-bean-high}}c}{Soy-bean-high}  & 324   & 1041  & 53.04  & \textbf{100.00} & \textbf{100.00} & 43.34  & 70.83  & 91.83  & \textbf{100.00} & 44.13  \\
    \multicolumn{1}{>{\columncolor{Soy-bean-mid}}c}{Soy-bean-mid} & 2548  & 35317 & 97.84  & 93.08  & 92.56  & 93.06  & \textbf{100.00} & 60.61  & 91.55  & 99.91  \\
    \multicolumn{1}{>{\columncolor{Soy-bean-low}}c}{Soy-bean-low} & 1428  & 27782 & 88.28  & \textbf{99.72} & 89.08  & 75.26  & 92.41  & 82.93  & 79.89  & 91.90  \\
    \multicolumn{1}{>{\columncolor{Residues}}c}{Residues} & 368   & 5427  & 94.97  & 84.48  & 73.01  & 77.92  & 90.60  & 42.95  & 72.18  & \textbf{100.00} \\
    \multicolumn{1}{>{\columncolor{Wheat}}c}{Wheat} & 182   & 3205  & 69.74  & 22.84  & 23.12  & 86.58  & 78.07  & 20.03  & 52.51  & \textbf{97.66}  \\
    \multicolumn{1}{>{\columncolor{Hay}}c}{Hay}   & 1938  & 48107 & 99.26  & 87.74  & 74.08  & 85.33  & 96.36  & 84.00  & 82.47  & \textbf{99.89} \\
    \multicolumn{1}{>{\columncolor{Grass/Pasture}}c}{Grass/Pasture} & 496   & 5048  & 85.72  & 97.54  & \textbf{100.00} & 68.16  & 84.53  & 60.24  & 98.75  & 99.98  \\
    \multicolumn{1}{>{\columncolor{Cover Crop 1}}c}{Cover Crop 1} & 400   & 2346  & 25.66  & 98.76  & 98.08  & 8.06  & 37.55  & 71.27  & \textbf{99.23} & 29.64  \\
    \multicolumn{1}{>{\columncolor{Cover Crop 2}}c}{Cover Crop 2} & 176   & 1988  & 99.64  & \textbf{100.00} & \textbf{100.00} & \textbf{100.00} & \textbf{100.00} & 87.93 & \textbf{100.00} & \textbf{100.00} \\
     \multicolumn{1}{>{\columncolor{Woodlands}}c}{Woodlands} & 1640  & 46919 & 87.59  & 95.20  & 73.35  & 68.60  & 97.93  & 91.36 & \textbf{99.62}  & 94.26  \\
    \multicolumn{1}{>{\columncolor{Highway}}c}{Highway} & 105   & 4758  & 98.17  & 96.51  & 98.51  & \textbf{100.00}  & 92.06  & 86.21  & \textbf{100.00} & 97.24  \\
    \multicolumn{1}{>{\columncolor{Local Road}}c}{Local Road} & 52    & 450   & 25.80  & 92.44  & 48.00  & 8.79  & 57.44  & 97.78  & \textbf{99.78} & 96.57  \\
    \multicolumn{1}{>{\columncolor{Buildings}}c}{Buildings} & 40    & 506   & 56.94  & 17.19  & 0.00  & \textbf{100.00} & 99.21  & 7.11  & 2.17  & \textbf{100.00} \\\midrule
    \multicolumn{3}{c}{OA} & 88.04  & 92.10  & 81.07  & 67.61  & 87.90  & 78.26  & 88.89  & \textbf{93.27} \\
    \multicolumn{3}{c}{AA} & 75.35  & 86.38  & 79.36  & 73.58  & 76.04  & 72.55  & 86.13  & \textbf{90.31} \\
    \multicolumn{3}{c}{Kappa} & 85.36  & 90.37  & 77.52  & 61.93  & 85.29  & 73.64  & 86.54  & \textbf{91.74} \\\bottomrule
    \end{tabular}%
  \label{tab:India2010}%
\end{table*}%

\begin{figure}[!tp]
\centering
\centerline{\subfigure[OA=71.24\%]{\includegraphics[scale=0.2]{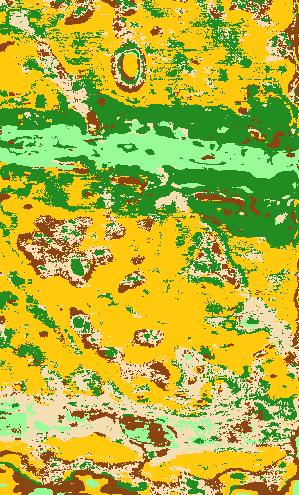}}
			\subfigure[OA=76.19\%]{\includegraphics[scale=0.2]{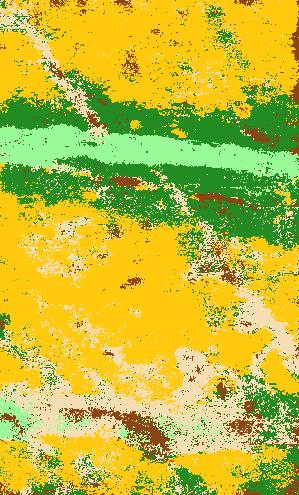}}
            \subfigure[OA=80.53\%]{\includegraphics[scale=0.2]{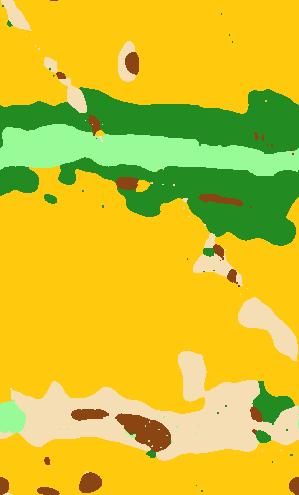}}
            \subfigure[OA=79.31\%]{\includegraphics[scale=0.2]{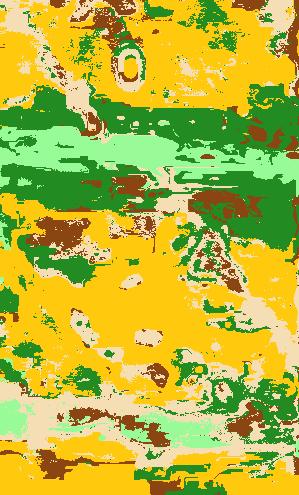}}}
\centerline{\subfigure[OA=80.19\%]{\includegraphics[scale=0.2]{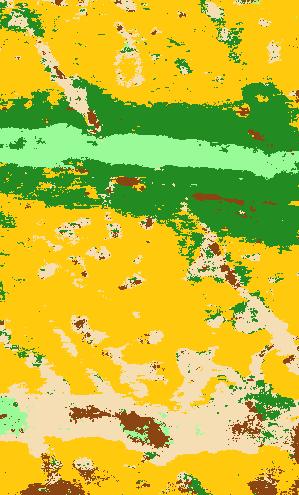}}
			\subfigure[OA=84.80\%]{\includegraphics[scale=0.2]{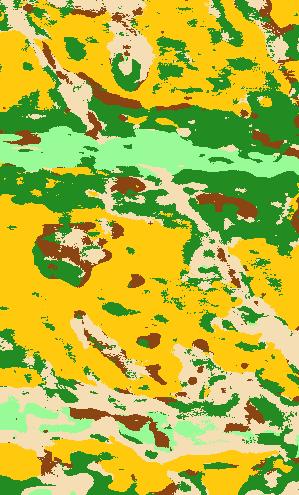}}
            \subfigure[OA=74.17\%]{\includegraphics[scale=0.2]{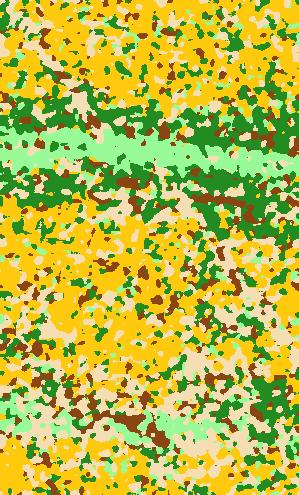}}
			\subfigure[OA=92.46\%]{\includegraphics[scale=0.2]{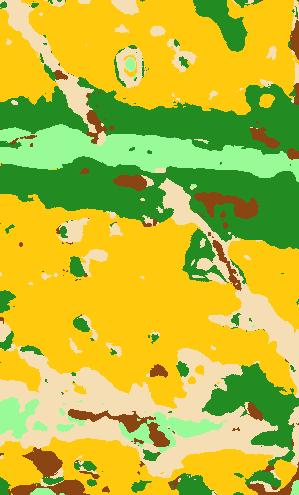}}}
\caption{Classification maps of all studied approaches on the TS4-1900 data set. (a) SVM \cite{SVM}, (b) MFL \cite{MFL}, (c)  WMRF \cite{WMRF}, (d) SCMK \cite{SCMK}, (e) OTVCA \cite{OTVCA}, (f) RPNet \cite{XU2018344}, (g) GTR \cite{GTR}, and (h) Our method. }
\label{TSresult}
\end{figure}
\subsection{Classification Results}
1) Crop Scene\par
The first experiment is conducted on the crop scene, i.e., Indian Pines 2010 data set. Fig. \ref{Indiaresult} gives the classification maps of all studied approaches. As shown in Fig. \ref{Indiaresult}, the spectral classifier, i.e., SVM, produces mis-classified noisy labels in the visual map. This is mainly because the SVM method only utilizes spectral information without including the spatial information. The MFL method is slightly superior to the SVM method in improving the mislabels. However, there are still noisy spots around the edges among different objects. The OTVCA, RPNet, and GTR methods improve the \lq\lq salt and pepper\rq\rq\ appearance in some homogenous areas because these methods integrate spatial and spectral information. The WMRF method yields an over-smoothed classification map since the training set is used as an additional constraint in the Markov model. The SCMK method fails to accurately identify the land covers with large regions, such as Cover Crop 1 and Local Road (corresponding to the eleventh and fifteenth classes). The main reason is that this image contains objects with different sizes. It is difficult to select the optimal number of superpixels. By contrast, the proposed method produces a promising classification map by integrating both pre-processing spatial features and post-processing spatial optimization. It can be seen from Fig. \ref{Indiaresult} (h) that the classification map contains fewer noisy spots than other approaches. In addition, the OA of each method is also shown in Fig. \ref{Indiaresult}. It can be observed that our method obtains the highest OA over other compared approaches.\par

Table \ref{tab:India2010} presents the quantitative results of all considered approaches. It can be observed that the individual accuracy of the Buildings class obtained by the WMRF method is zero. This is due to the fact that the spatial constraint in the WMRF model smooths out the small-sized objects. Among all studied approaches, our method yields the highest classification accuracies in terms of OA, AA, and Kappa coefficient, which also further validates that the proposed method can better achieve the classification of HSI in crop scene.

\begin{table*}[!htbp]
  \centering
  \caption{Classification Performance of All Studied Approaches for TS4-1900 Data Set, Including SVM \cite{SVM}, MFL \cite{MFL}, WMRF \cite{WMRF}, SCMK \cite{SCMK}, OTVCA \cite{OTVCA}, RPNet \cite{XU2018344}, GTR \cite{GTR} and Our Method.}
    \begin{tabular}{ccccccccccc}\toprule
    \multirow{2}[0]{*}[-2pt]{Class name} & \multirow{2}[0]{*}[-2pt]{Training set} & \multirow{2}[0]{*}[-2pt]{Testing set} & \multicolumn{8}{c}{Classification accuracies of different approaches (in \%)} \\\cmidrule(lr){4-11}
          &       &       & SVM   & MFL   & WMRF  & SCMK  & OTVCA & RPNet  & GTR   & Our method \\\midrule
    \multicolumn{1}{>{\columncolor{Anhydrite}}c}{Anhydrite} & 240   & 1795  & 85.30  & 76.77  & 77.27  & 83.23  & \textbf{92.55} & 90.03  & 83.18  & 92.26  \\
    \multicolumn{1}{>{\columncolor{Quartz}}c}{Quartz} & 317   & 2467  & 68.62  & 64.33  & 65.30  & 74.49  & 72.67  & 83.18  & 66.19  & \textbf{89.00} \\
    \multicolumn{1}{>{\columncolor{Sulfides}}c}{Sulfides}& 245   & 676   & 35.83  & 65.53  & 78.70  & 46.14  & 57.14  & \textbf{95.56}  & 95.12 & 77.75  \\
     \multicolumn{1}{>{\columncolor{Muscovite}}c}{Muscovite} & 403   & 2998  & 70.76  & 77.25  & 74.65  & 82.79  & 77.60  & 83.19  & 68.91  & \textbf{94.07} \\
    \multicolumn{1}{>{\columncolor{Feldspar}}c}{Feldspar} & 335   & 3366  & 80.72  & 85.77  & \textbf{99.05} & 91.79  & 87.38  & 82.47  & 75.70  & 97.42  \\\midrule
    \multicolumn{3}{c}{OA} & 71.24  & 76.19  & 80.53  & 79.31  & 80.19  & 84.80  & 74.17  & \textbf{92.46} \\
    \multicolumn{3}{c}{AA} & 68.25  & 73.93  & 78.99  & 75.69  & 77.46  & 86.89  & 77.82  & \textbf{90.10} \\
    \multicolumn{3}{c}{Kappa} & 62.84  & 69.09  & 74.33  & 73.36  & 74.04  & 80.22  & 66.84  & \textbf{90.18} \\\bottomrule
    \end{tabular}%
  \label{tab:TS1900}%
\end{table*}%

2) Mineral scene\par
The second experiment is conducted on the mineral scene. The mineral maps of all studied approaches are shown in Fig. \ref{TSresult}. It can be easily seen that the SVM, which only utilizes the spectral information as input, still tends to produce a noisy estimation in the resulting map. The MFL method also yields noise-like mislabeled pixels in the classification result. For the WMRF method, the classification map suffers from the over-smoothed phenomenon. The GTR method also cannot work well in mineral mapping, since the tensor regression model cannot well fit the spectral curves of different minerals with low discrimination. In contrast, the proposed method is able to produce high-quality classification results due to its effective performance in dealing with the low discrimination of different classes. From the OA listed in Fig. \ref{TSresult}, it is shown that our method obtains the highest value for mineral mapping.\par
The objective metrics in terms of classification accuracy are presented in Table \ref{tab:TS1900}. It can be observed that the proposed method outperforms other classification methods in terms of three objective indexes, with OA=92.46\%, AA=90.10\%, and Kappa=90.18\%. This also confirms the effectiveness of our method for mineral mapping.\par
\begin{figure*}[!tp]
\centering
\centerline{\subfigure[OA=74.47\%]{\includegraphics[scale=0.12]{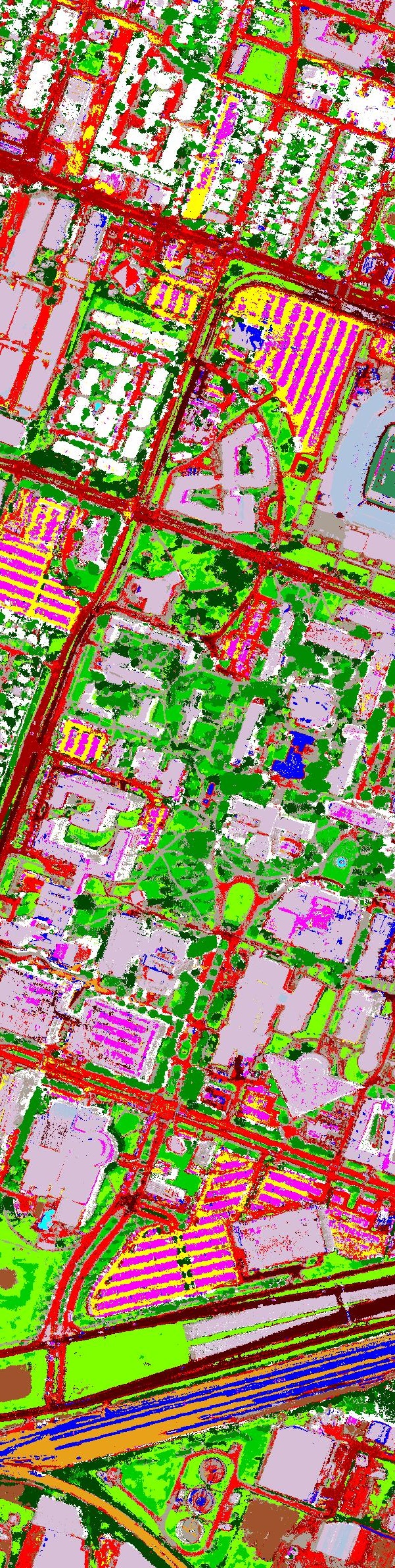}}
			\subfigure[OA=66.44\%]{\includegraphics[scale=0.12]{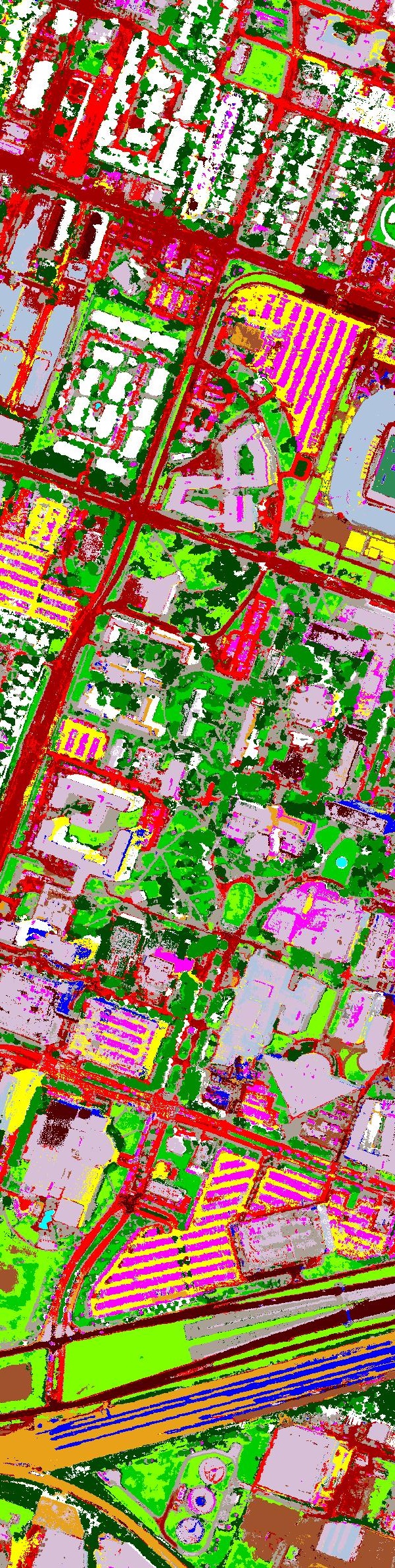}}
            \subfigure[OA=78.22\%]{\includegraphics[scale=0.12]{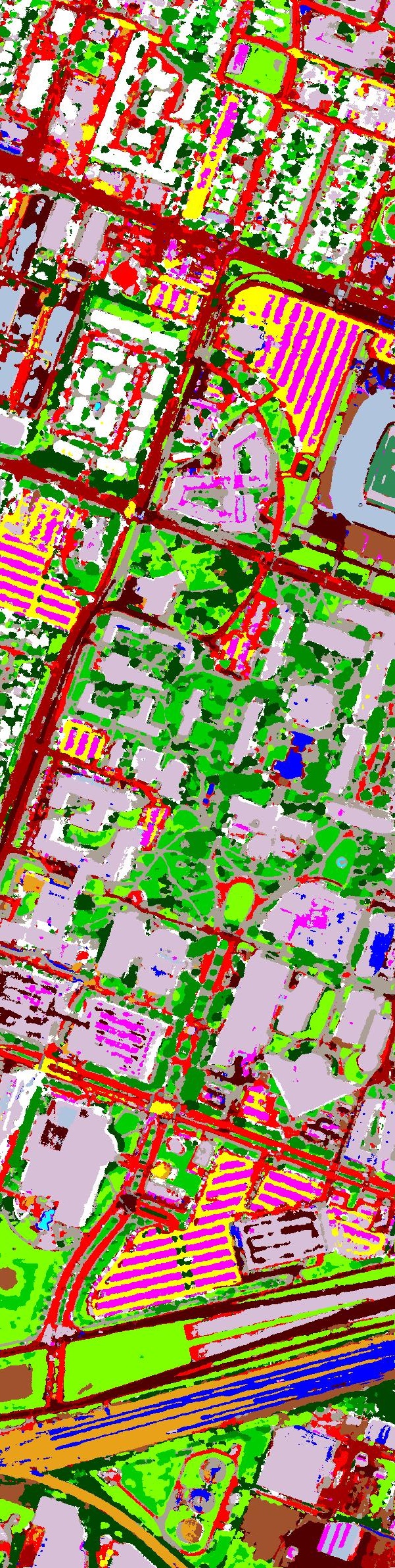}}
            \subfigure[OA=80.67\%]{\includegraphics[scale=0.12]{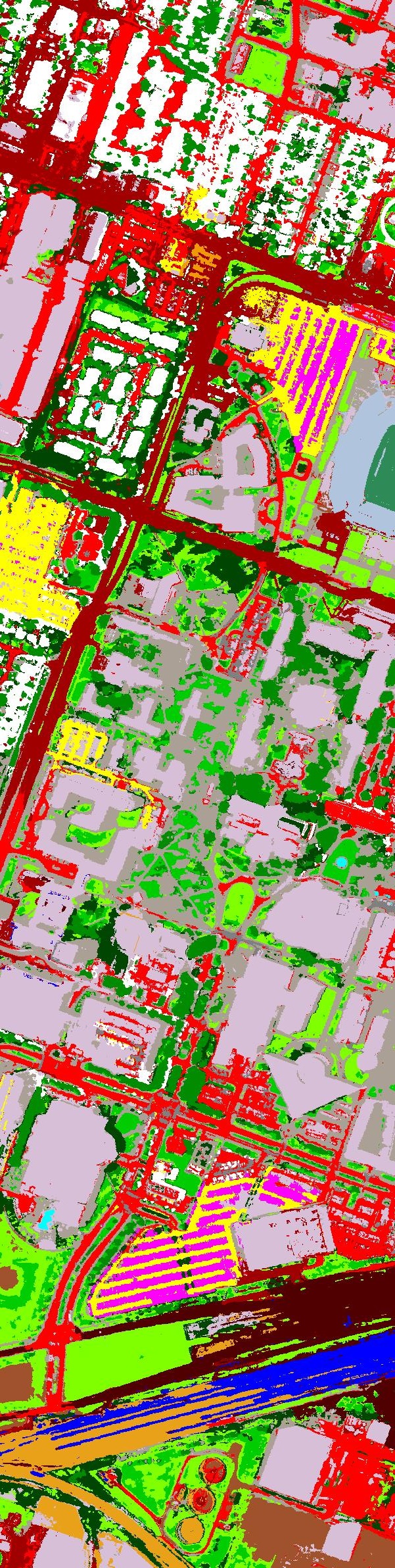}}
            \subfigure[OA=88.08\%]{\includegraphics[scale=0.12]{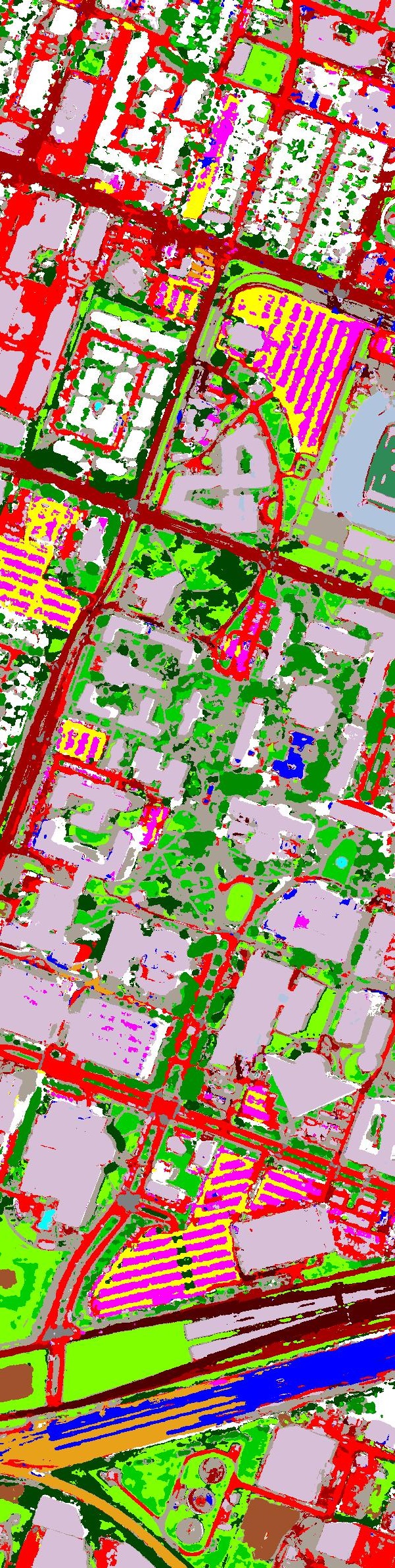}}
			\subfigure[OA=75.36\%]{\includegraphics[scale=0.12]{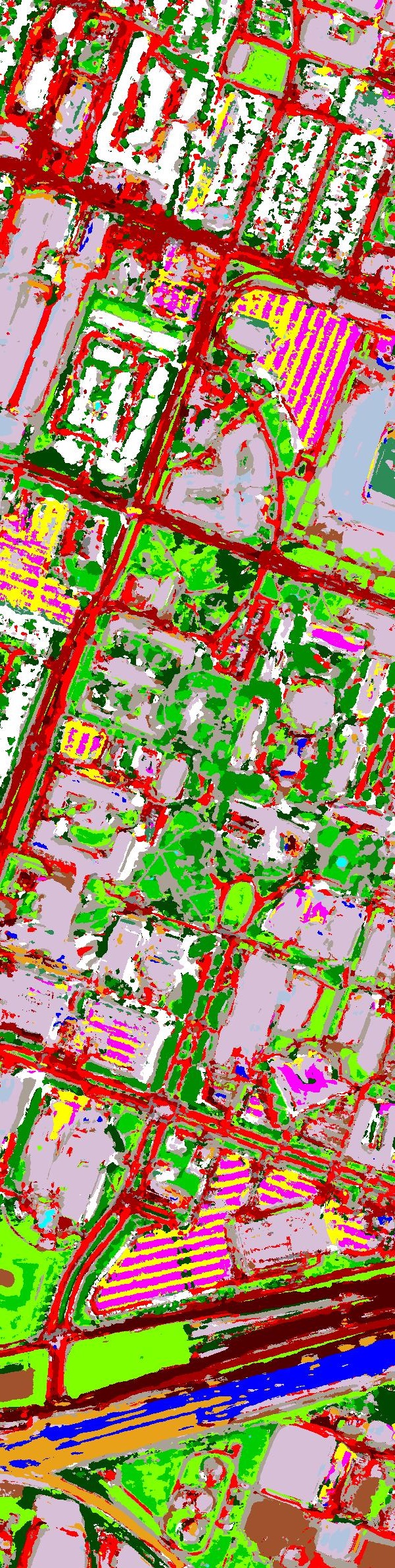}}
            \subfigure[OA=54.87\%]{\includegraphics[scale=0.12]{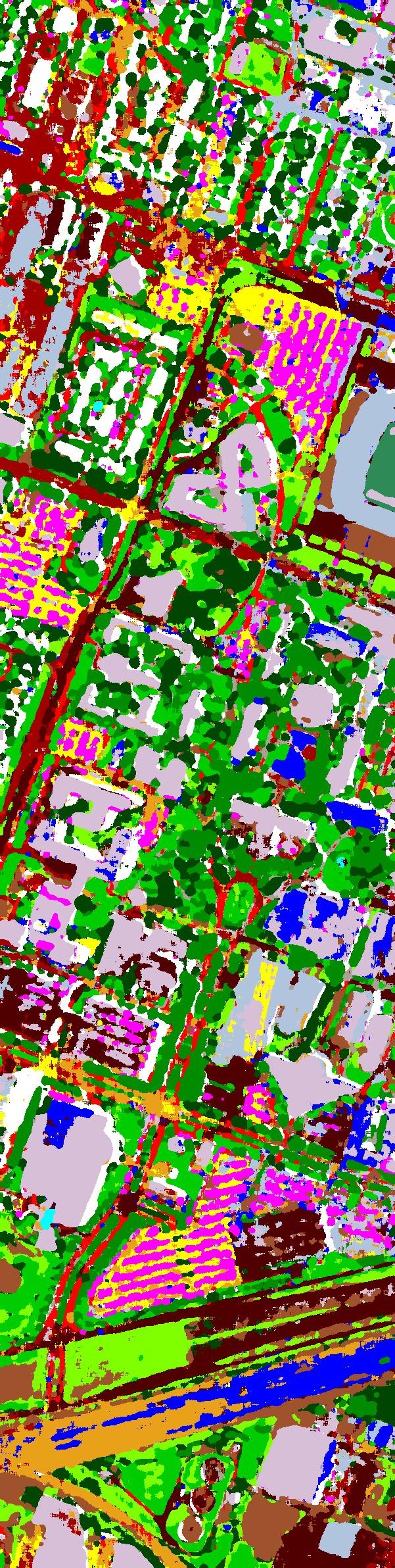}}
			\subfigure[OA=91.73\%]{\includegraphics[scale=0.12]{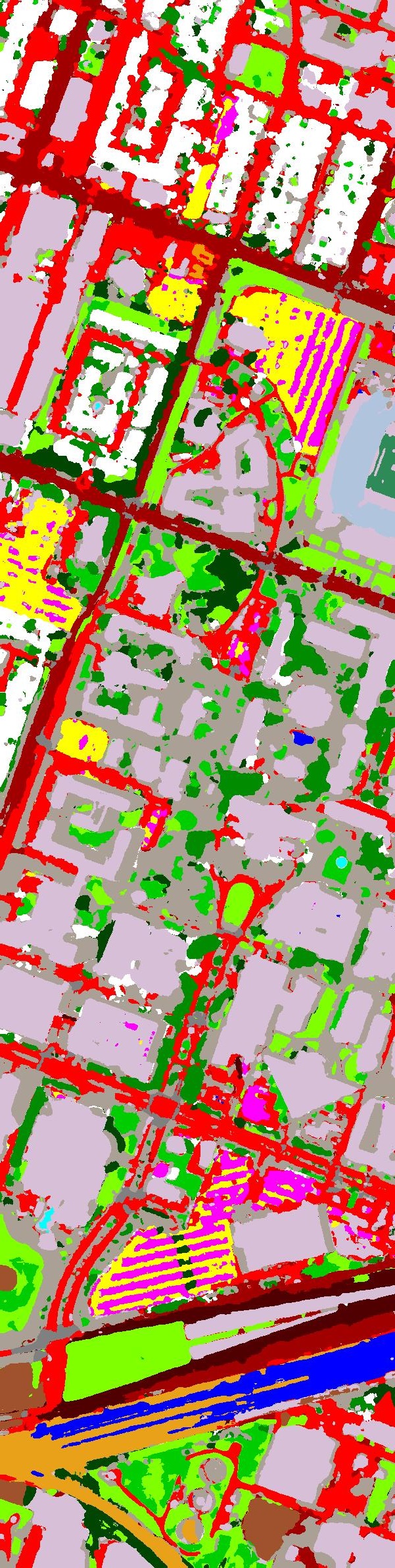}}}
\caption{Classification maps of all studied approaches on the Houston2018 data set. (a) SVM \cite{SVM}, (b) MFL \cite{MFL}, (c) WMRF \cite{WMRF}, (d) SCMK \cite{SCMK}, (e) OTVCA \cite{OTVCA}, (f) RPNet \cite{XU2018344}, (g) GTR \cite{GTR}, and (h) Our method. }
\label{Huresult}
\end{figure*}
\begin{table*}[htbp]
  \centering
  \caption{Classification Performance of All Studied Approaches for Houston 2018 Data Set, Including SVM \cite{SVM}, MFL \cite{MFL}, WMRF \cite{WMRF}, SCMK \cite{SCMK}, OTVCA \cite{OTVCA}, RPNet \cite{XU2018344}, GTR \cite{GTR} and Our Method.}
    \begin{tabular}{ccccccccccc}\toprule
    \multirow{2}[0]{*}[-2pt]{Class name} & \multirow{2}[0]{*}[-2pt]{Training set} & \multirow{2}[0]{*}[-2pt]{Testing set} & \multicolumn{8}{c}{Classification accuracies of different approaches (in \%)} \\\cmidrule(lr){4-11}
          &       &       & SVM   & MFL   & WMRF  & SCMK  & OTVCA & RPNet  & GTR   & Our method \\\midrule
    \multicolumn{1}{>{\columncolor{Healthy Grass}}c}{Healthy Grass} & 500   & 9299  & 88.44  & 89.85  & 94.44 & 78.38  & 75.14  & \textbf{96.41}  & 92.49  & 68.27  \\
    \multicolumn{1}{>{\columncolor{Stressed Grass}}c}{Stressed Grass} & 500   & 32002 & 90.44  & 84.57  & 93.06  & 90.25  & \textbf{93.34}  & 90.15  & 59.87  & 92.81 \\
    \multicolumn{1}{>{\columncolor{Artificial Turf}}c}{Artificial Turf} & 68    & 616   & 98.56  & \textbf{100.00} & \textbf{100.00} & \textbf{100.00} & 99.68  & 96.75  & 99.68  & \textbf{100.00} \\
    \multicolumn{1}{>{\columncolor{Evergreen Trees}}c}{Evergreen Trees} & 500   & 13088 & 80.09  & 92.96  & \textbf{97.94}  & 81.13  & 82.25  & 96.85 & 96.67  & 85.85  \\
    \multicolumn{1}{>{\columncolor{Deciduous Trees}}c}{Deciduous Trees} & 500   & 4548  & 43.04  & 91.38  & \textbf{95.89}  & 35.35  & 62.03  & 94.85 & 91.71  & 78.48 \\
    \multicolumn{1}{>{\columncolor{Bare Earth}}c}{Bare Earth} & 451   & 4065  & 73.83  & 97.37  & \textbf{100.00} & 79.69  & 94.95  & 99.95  & \textbf{100.00} & 99.98  \\
    \multicolumn{1}{>{\columncolor{Water}}c}{Water} & 26    & 240   & \textbf{100.00}  & 98.75  & 92.50  & 85.61  & \textbf{100.00} & 95.00  & 90.42  & \textbf{100.00}  \\
    \multicolumn{1}{>{\columncolor{Residential}}c}{Residential}  & 500   & 39262 & 58.85  & 81.03  & 84.09  & 78.85  & 79.17  & 86.99  & 72.15  & \textbf{93.42} \\
    \multicolumn{1}{>{\columncolor{Commercial}}c}{Commercial} & 500   & 223184 & 97.38  & 60.24  & 79.78  & 97.80  & \textbf{98.59}  & 71.66  & 55.86  & 98.54  \\
    \multicolumn{1}{>{\columncolor{Roads}}c}{Roads} & 500   & 45310 & 57.04  & 47.02  & 54.61  & 59.30  & 82.07  & 57.85  & 25.74  & \textbf{86.34} \\
    \multicolumn{1}{>{\columncolor{Sidewalks}}c}{Sidewalks} & 500   & 33502 & 54.36  & 47.61  & 58.43  & 58.11  & 66.27  & 52.38  & 8.19  & \textbf{66.60} \\
    \multicolumn{1}{>{\columncolor{Crosswalks}}c}{Crosswalks} & 151   & 1365  & 23.01  & 11.65  & 38.32  & 10.60  & 38.28  & \textbf{41.90} & 0.00  & 35.02  \\
    \multicolumn{1}{>{\columncolor{Major Thoroughfares}}c}{Major Thoroughfares} & 500   & 45858 & 64.22  & 62.89  & 63.95  & 68.86  & 84.95  & 72.64  & 40.09  & \textbf{92.65} \\
    \multicolumn{1}{>{\columncolor{Highways}}c}{Highways} & 500   & 9349  & 46.95  & 88.30  & 92.94  & 58.12  & 82.20  & \textbf{98.73} & 94.16  & 87.29  \\
    \multicolumn{1}{>{\columncolor{Railways}}c}{Railways} & 500   & 6437  & 83.06  & 98.70  & \textbf{99.91}  & 69.03  & 93.21  & 98.85 & 95.57  & 99.29  \\
    \multicolumn{1}{>{\columncolor{Paved Parking Lots}}c}{Paved Parking Lots} & 500   & 10975 & 70.68  & 93.31  & \textbf{95.83} & 81.19  & 93.30  & 94.10  & 64.39  & 94.87  \\
    \multicolumn{1}{>{\columncolor{Gravel Parking Lots}}c}{Gravel Parking Lots} & 14    & 135   & 56.47  & 43.70  & 15.56  & 85.99  & \textbf{97.12} & 95.56  & 0.00  & \textbf{99.26} \\
    \multicolumn{1}{>{\columncolor{Cars}}c}{Cars}  & 500   & 6078  & 29.56  & 89.47  & \textbf{94.75}  & 87.80  & 63.75  & 92.00 & 85.70  & 92.21  \\
    \multicolumn{1}{>{\columncolor{Trains}}c}{Trains} & 500   & 4865  & 42.03  & 95.29  & 97.57  & 93.66  & 68.87  & 99.24 & 74.98  & \textbf{99.46}  \\
    \multicolumn{1}{>{\columncolor{Stadium Seats}}c}{Stadium Seats} & 500   & 6324  & 71.57  & 91.89  & 99.38  & 99.79  & 92.70  & 99.56  & 99.76  & \textbf{99.97} \\\midrule
    \multicolumn{3}{c}{OA} & 74.47  & 66.44  & 78.22  & 80.67  & 88.08  & 75.36  & 54.87  & \textbf{91.73} \\
    \multicolumn{3}{c}{AA} & 66.48  & 78.30  & 82.45  & 74.97  & 82.39  & 86.57 & 67.37  & \textbf{88.52}  \\
    \multicolumn{3}{c}{Kappa} & 68.39  & 59.87  & 72.77  & 75.72  & 84.68  & 69.70  & 47.23  & \textbf{89.28} \\\bottomrule
    \end{tabular}%
  \label{tab:hu18}%
\end{table*}%

\begin{table*}[!tbp]
  \centering
  \caption{The Computing Time of Different Approaches.}
    \begin{tabular}{ccccccccc}\toprule
    \multirow{2}[0]{*}[-2pt]{Data sets} & \multicolumn{8}{c}{Running time of all studied approaches (in seconds)} \\\cmidrule(lr){2-9}
          & SVM  \cite{SVM}  & MFL \cite{MFL}  & WMRF \cite{WMRF}  & SCMK \cite{SCMK}  & OTVCA \cite{OTVCA} & RPNet \cite{XU2018344} & GTR \cite{GTR}  & Our method \\\midrule
    Indian Pines 2010 & 3009.32 & \textbf{64.05} & 1877.82 & 996.25  & 684.41  & 174.43 & 165.96 & 1034.62 \\
    TS4-1900 & 1243.02 & \textbf{89.48} & 219.31 & 116.78 & 479.18 & 92.95 & 125.85 & 221.44 \\
    Houston 2018 & 1743.71 & \textbf{174.96} & 1758.42 & 1989.85 & 2095.41 & 1111.25 & 323.31 & 2034.06 \\\bottomrule
    \end{tabular}%
  \label{tab:time}%
\end{table*}%
3) Urban scene\par
To examine the performance of our method for classification of urban areas, the third experiment is performed on a challenging data set, i.e., Houston 2018. Fig. \ref{Huresult} provides the classification maps of different approaches. As shown in Fig. \ref{Huresult}, the SVM and MFL approaches still obtain different levels of \lq\lq noise\rq\rq\ like mislabels (see the Commercial class in Fig. \ref{Huresult}). By introducing spatial constraints, the WMRF method improves the \lq\lq noise \rq\rq\ phenomenon. The GTR method suffers from serious misclassification in the classification map, especially for Sidewalks and Crosswalks classes. By contrast, our method yields a better classification result, in which the boundaries of different objects are well aligned with the labeled Ground Truth. The reason is twofold. One is that the feature extraction technique well enhances the separabilities belonging to different classes. Another is that the complementary characteristics from feature extraction and spatial optimization are sufficiently exploited in the proposed method. It should be mentioned that the CA of Healthy Grass is relatively low compared to other methods. The reason is that our method fails to distinguish between Healthy Grass and Stressed Grass well due to the similarity in their spectral curves. According to the OA presented in Fig. \ref{Huresult}, it is easy to observe that our method still provides competitive OA than other studied approaches.\par
Besides, the classification accuracies of all studied approaches are reported in Table \ref{tab:hu18}. The proposed method produces the highest OA, AA, and Kappa coefficient compared to other competitive approaches. This is similar to the previous results on the Indian Pines 2010 and TS4-1900 data sets. In addition, our method yields the highest individual accuracies for nine different classes. It can be concluded that the proposed method can work well in classifying different land covers including crop areas, mineral areas, and urban areas.


\subsection{Computing Cost}
In this work, all experiments are implemented with Matlab 2014a on a Laptop with Intel(R) Core(TM) i7-6700HQ CPU processor (2.6 GHz), 8GB of memory and 64-bit Operating System. The running time of all studied approaches on three used data sets is given in Table \ref{tab:time}. It can be observed that the computing time tends to linearly increase as the spatial size and the spectral dimension of data increases. In addition, the computing time of our method is acceptable. Taking the TS4-1900 image as an example, the computing time of the proposed method for mineral mapping is about 221.44 s. The MFL method is the fastest because this method adopts the MLR classifier without cross-validation like SVM. How to reduce the processing time is an interesting topic that is not within the scope of this paper.

\section{Conclusions}
\label{sec:cons}
In this work, a novel classification framework for hyperspectral images is proposed based on the fusion of dual spatial information. Here, the pre-processing spatial feature extraction and post-processing spatial optimization are considered. First, the spectral dimension of the original image is decreased with an averaging-based scheme. Then, we develop an SP to smooth out the unimportant component of the dimension-reduced data followed by the spectral classifier. Next, the SVM classifier is performed on the dimension-reduced data to obtain the initial probability, and the ERW is exploited to refine the initial probability by considering the relationship among neighboring pixels. Finally, both levels of class probabilities are integrated together via the decision fusion rule to yield the final classification map. Through experimental analysis on three imaging scenes, several conclusions can be summarized:\par
1) In contrast to other widely used feature extraction methods, the proposed SP provides obvious advantages in decreasing the spectral variability of pixels belonging to the same class and increasing the discrimination of different land covers, which examines the superiority and potential of the SP for classification.\par
2) The proposed classification framework always yields higher performance when the feature extraction and spatial optimization techniques are merged. This is due to that the proposed framework flexibly merges both levels of spatial information. \par
3) Experiments performed on three hyperspectral data sets from different scenes confirm that the proposed method produces competitive classification performance over other compared approaches, which also illustrates the generalization capability and effectiveness of the proposed method for identifying different minerals and materials.\par
In the future, how to automatically obtain the parameters in the feature extraction process will be investigated.
\section*{Acknowledgement}
The authors would like to thank Prof. Melba Crawford for sharing the Indian Pines 2010 data set, and Dr. Peter Seidel for preparing the TS4-1900 data set, and the IEEE GRSS Image Analysis and Data Fusion Technical Committee for distributing the Houston 2018 data set used in this paper.

%

\appendices
\ifCLASSOPTIONcaptionsoff
  \newpage
\fi


\bibliographystyle{IEEEtran}
\bibliography{IEEEabrv,refs}
\begin{IEEEbiography}[{\includegraphics[width=1in,height=1.25in]{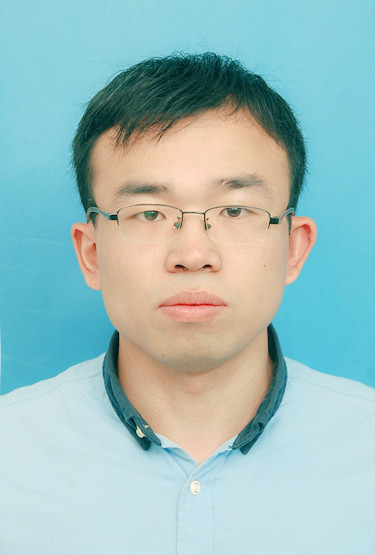}}]{Puhong Duan}
(S'17) received the B.Sc degree from Suzhou University, Suzhou, China, in 2014, and the M.S. degree from Hefei University of Technology, Hefei, China, in 2017. He is currently working toward the Ph.D. degree in the Laboratory of Vision and Image Processing, Hunan University, Changsha, China.\par
From October 2019 to December 2019, he was a Visiting Ph. D. Student with the Faculty of Electrical Engineering and Computer Science, Technische Universität Berlin. From January 2020 to October 2020, he was a Visiting Ph. D. Student with the Helmholtz-Zentrum Dresden-Rossendorf, Helmholtz Institute Freiberg for Resource Technology. His research interests include hyperspectral image classification, visualization, and image fusion.
\end{IEEEbiography}

\begin{IEEEbiography}[{\includegraphics[width=1in,height=1.25in]{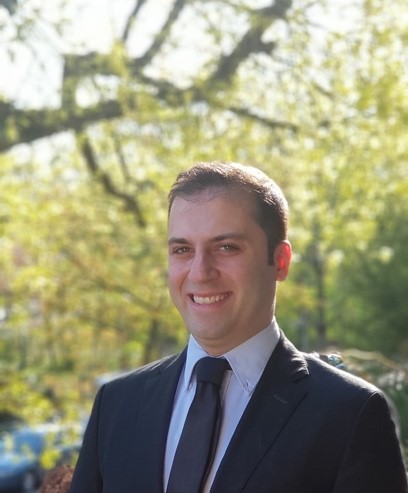}}]{Pedram Ghamisi}(S'12-M'15-SM'18)
received the B.Sc. degree in civil (survey) engineering from the Tehran South Campus of Azad University, Tehran, Iran, the M.Sc. degree (First Class Hons.) in remote sensing from the K. N. Toosi University of Technology, Tehran, in 2012, and the Ph.D. degree in electrical and computer engineering from the University of Iceland, Reykjavik, Iceland, in 2015. \par
He works as the head of the machine learning group at Helmholtz-Zentrum Dresden-Rossendorf (HZDR), Germany and as the CTO and co-founder of VasoGnosis Inc, USA. He is also the co-chair of IEEE Image Analysis and Data Fusion Committee. Dr. Ghamisi was a recipient of the Best Researcher Award for M.Sc. students at the K. N. Toosi University of Technology in the academic year of 2010-2011, the IEEE Mikio Takagi Prize for winning the Student Paper Competition at IEEE International Geoscience and Remote Sensing Symposium (IGARSS) in 2013, the Talented International Researcher by Iran’s National Elites Foundation in 2016, the first prize of the data fusion contest organized by the Image Analysis and Data Fusion Technical Committee (IADF) of IEEE-GRSS in 2017, the Best Reviewer Prize of IEEE Geoscience and Remote Sensing Letters (GRSL) in 2017, the IEEE Geoscience and Remote Sensing Society 2020 Highest Impact Paper Award, the Alexander von Humboldt Fellowship from the Technical University of Munich, and the High Potential Program Award from HZDR. He serves as an Associate Editor for MDPI-Remote Sensing, MDPI-Sensor, and IEEE GRSL. His research interests include interdisciplinary research on machine (deep) learning, image and signal processing, and multisensor data fusion. For detailed info, please see http://pedram-ghamisi.com/.
\end{IEEEbiography}

\begin{IEEEbiography}[{\includegraphics[width=1in,height=1.25in]{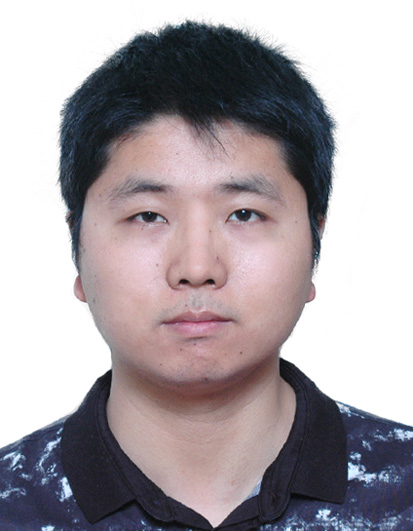}}]{Xudong Kang}
(S'13-M'15-SM'17) received the B.Sc degree from Northeast University, Shenyang, China, in 2007, and the Ph.D. degree from Hunan University, Changsha, China, in 2015.\par
In 2015, he joined the college of electrical engineering of Hunan University, Changsha, China. His research interest includes hyperspectral feature extraction, image classification, image fusion, and anomaly detection. \par
Dr. Kang received the National Nature Science Award of China (Second class and Rank as third) and the Second Prize in the Student Paper Competition in IGARSS 2014. He was also selected as the Best Reviewer for the IEEE GEOSCIENCE AND REMOTE SENSING LETTERS and the IEEE TRANSACTIONS ON GEOSCIENCE AND REMOTE SENSING. He was an Associate Editor of the IEEE TRANSACTIONS ON GEOSCIENCE AND REMOTE SENSING from 2018 to 2019. He serves as an Associate Editor for the IEEE GEOSCIENCE AND REMOTE SENSING LETTERS and the IEEE JOURNAL ON MINIATURIZATION FOR AIR AND SPACE SYSTEMS.
\end{IEEEbiography}

\begin{IEEEbiography}[{\includegraphics[width=1in,height=1.25in]{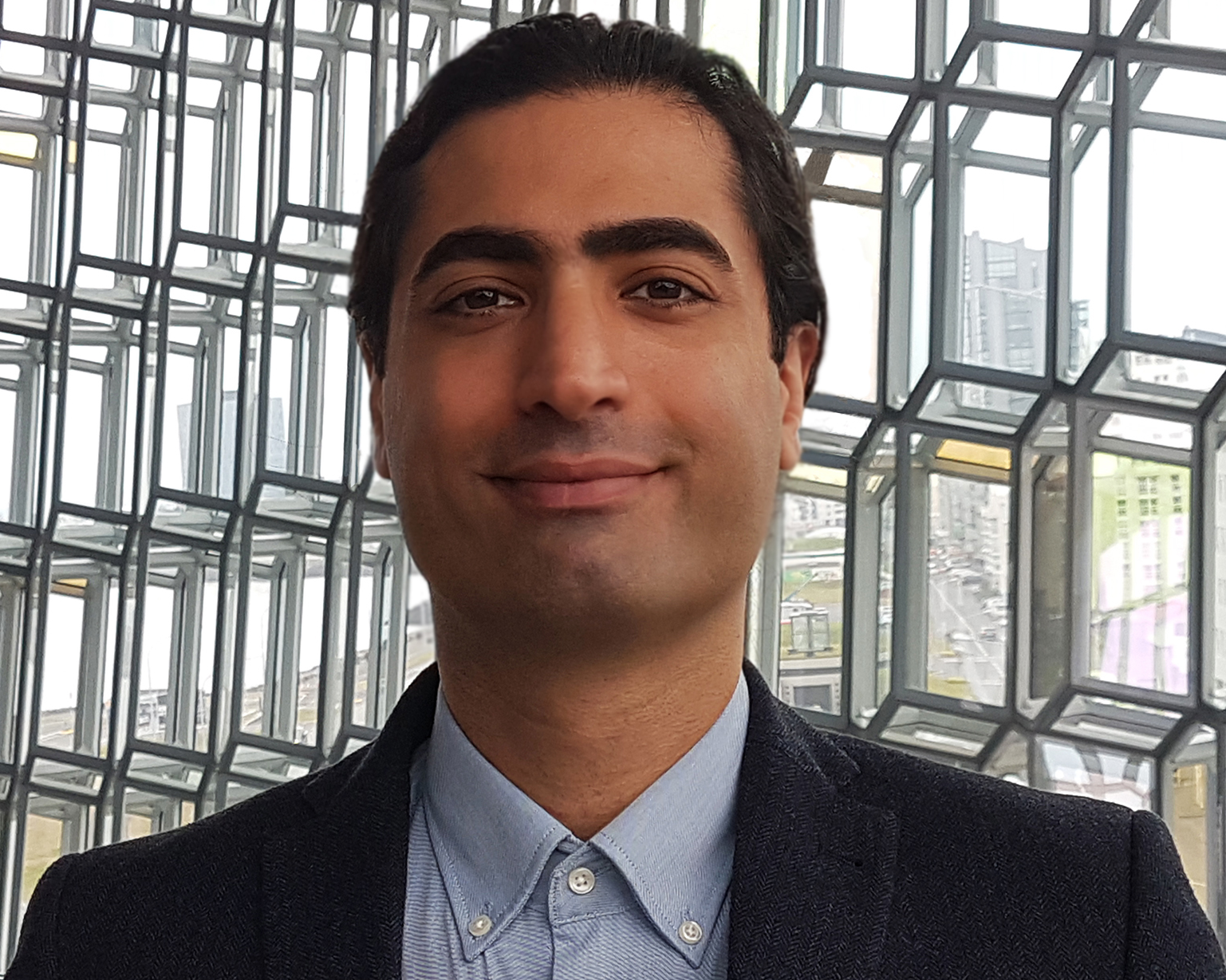}}]{Behnood Rasti}
 received the B.Sc. and M.Sc. degrees both in Electronics-Electrical Engineering from the Electrical Engineering Department, University of Guilan, Rasht, Iran, in 2006, and 2009, respectively, and received the Ph.D. degree in Electrical and Computer Engineering from the University of Iceland, Reykjavik, Iceland, in 2014. He was the valedictorian as an M.Sc. student in 2009 and he won the Doctoral Grant of The University of Iceland Research Fund and was awarded "The Eimskip University fund", in 2013.\par
 In 2015 and 2016, Dr. Rasti worked as a postdoctoral researcher and a seasonal lecturer at the Electrical and Computer Engineering Department, University of Iceland. From 2016 to 2019, he has been a lecturer at the Center of Engineering Technology, Department of Electrical and Computer Engineering, University of Iceland where he has been teaching several core courses such as Linear Systems, Control Systems, Sensors and Actuators, Data Acquisition and Processing, Circuit Theories, Electronics, and PLC programming. He has supervised several research projects and engineering B.Sc. theses as a lecturer.\par
 In 2019, Dr. Rasti won the prestigious “Alexander von Humboldt Research Fellowship Grant” and started his work in 2020 as a Humboldt research fellow with the Machine Learning Group, Helmholtz-Zentrum Dresden-Rossendorf (HZDR), Freiberg, Germany. His research interests include signal and image processing, machine/deep learning, remote sensing image fusion, hyperspectral feature extraction, hyperspectral unmixing, remote sensing image denoising and restoration.\par
Dr. Rasti serves as an Associate Editor for the IEEE GEOSCIENCE AND REMOTE SENSING LETTERS (GRSL) and Remote Sensing.
\end{IEEEbiography}
\begin{IEEEbiography}[{\includegraphics[width=1in,height=1.25in]{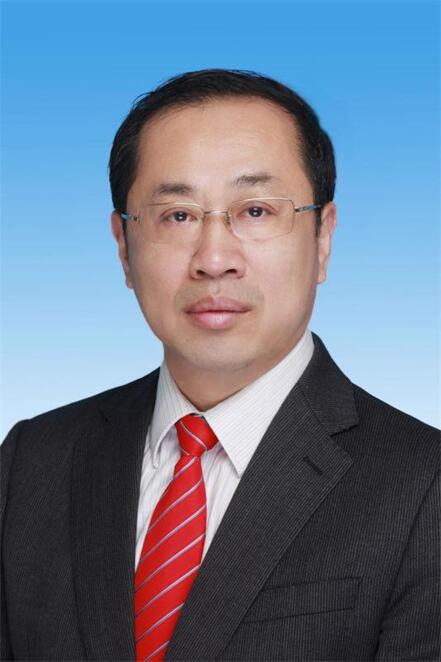}}]{Shutao Li}
(M'07–SM'15–F'19) received the B.S., M.S., and Ph.D. degrees from Hunan University, Changsha, China, in 1995, 1997, and 2001, respectively.\par
In 2001, he joined the College of Electrical and Information Engineering, Hunan University. From May 2001 to October 2001, He was a Research Associate with the Department of Computer Science, Hong Kong University of Science and Technology. From November 2002 to November 2003, he was a Postdoctoral Fellow with the Royal Holloway College, University of London. From April 2005 to June 2005, he was a Visiting Professor with the Department of Computer Science, Hong Kong University of Science and Technology. He is currently a Full Professor with the College of Electrical and Information Engineering, Hunan University. He has authored or co-authored over 200 refereed papers. He gained two 2nd-Grade State Scientific and Technological Progress Awards of China in 2004 and 2006.His current research interests include image processing, pattern recognition, and artificial intelligence. \par
He is an Associate Editor of the IEEE TRANSACTIONS ON GEOSCIENCE AND REMOTE SENSING and the IEEE TRANSACTIONS ON INSTRUMENTATION AND MEASUREMENT. He is a member of the Editorial Board of the Information Fusion and the Sensing and Imaging.
\end{IEEEbiography}

\begin{IEEEbiography}[{\includegraphics[width=1in,height=1.25in]{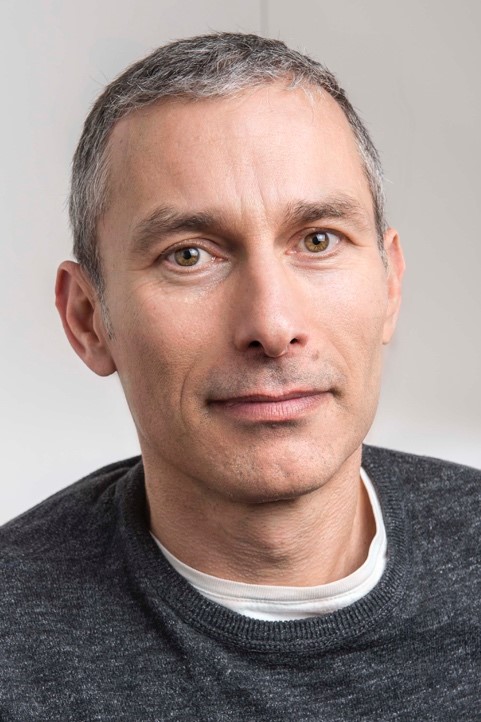}}]{Richard Gloaguen}
received the Ph.D. degree “Doctor Communitatis Europae” in marine geosciences from the University of Western Brittany, Brest, France, in collaboration with the Royal Holloway University of London, U.K., and Göttingen University, Germany, in 2000.\par
He was a Marie Curie Post-Doctoral Research Associate at the Royal Holloway University of London from 2000 to 2003. He led the Remote Sensing Group at University Bergakademie Freiberg, Freiberg, Germany, from 2003 to 2013. Since 2013, he has been leading the division “Exploration Technology” at the Helmholtz-Institute Freiberg for Resource Technology, Freiberg. He is currently involved in UAV-based multisource imaging, laser-induced fluorescence, and non-invasive exploration.\par
His research interests focus on multisource and multiscale remote sensing integration.\par
\end{IEEEbiography}
\end{document}